\documentclass{article} 
\usepackage{iclr2027_conference,times}

\usepackage{amsmath,amsfonts,bm}

\def\eqref#1{equation~\ref{#1}}

\def\1{\bm{1}}

\DeclareMathAlphabet{\mathsfit}{\encodingdefault}{\sfdefault}{m}{sl}
\SetMathAlphabet{\mathsfit}{bold}{\encodingdefault}{\sfdefault}{bx}{n}

\usepackage[T1]{fontenc}  
\usepackage{xcolor}
\definecolor{mycitecolor}{RGB}{71, 191, 38}
\definecolor{mylinkcolor}{RGB}{40, 115, 201}
\usepackage{hyperref}
\hypersetup{
  colorlinks=true,
  citecolor=mycitecolor,
  linkcolor=mylinkcolor,
  urlcolor=mycitecolor,
}
\usepackage{url}
\usepackage{amsmath,amssymb,amsfonts}
\usepackage{dsfont}          
\usepackage{graphicx}
\usepackage{booktabs}
\usepackage{multirow}
\usepackage[table]{xcolor}   
\usepackage{colortbl}
\usepackage{longtable}       
\usepackage{arydshln}        
\usepackage{makecell}        
\usepackage{subfig}          
\usepackage{wrapfig}         
\usepackage{tabularx}        
\usepackage{placeins}        
\usepackage{enumitem}
\usepackage{xspace}
\usepackage[most]{tcolorbox}
\usepackage[frozencache]{minted2}
\usepackage{mwe}

\usepackage{algorithm}
\usepackage{algpseudocode}
\usepackage{lipsum}

\usepackage{booktabs}
\usepackage{xltabular}

\newcommand{\popgen}{\textsc{PopGen}\xspace}
\newcommand{\popplay}{\textsc{PopPlay}\xspace}          

\newcommand{\environment}[1]{\texttt{#1}\xspace}

\title{
PopGen: Generating Semantically Diverse \\
Populations for Zero-Shot Coordination
}

\begingroup

\footnotetext{This work was conducted in part while the author was affiliated with Google DeepMind.}
\endgroup

\author{Paul Teiletche$^{1,2,*}$ \quad
Kanghyun Ryu$^{2,*}$ \quad
Kyungjoon Park$^{3}$ \quad
Dhruva Tirumala$^{4}$ \\[0.2em]
\bfseries
Markus Wulfmeier$^{5,\dagger}$ \quad
Negar Mehr$^{2}$ \\[0.5em]
\normalfont
$^{1}$EPFL \quad
$^{2}$UC Berkeley \quad
$^{3}$Yonsei University \quad
$^{4}$Google DeepMind \quad
$^{5}$Nomagic \\[0.2em]
$^{*}$Equal contribution 
}

\iclrfinalcopy 
\begin{document}

\maketitle
\lhead{} 

\begin{abstract}
In multi-agent reinforcement learning, exposing a cooperator to partners with diverse strategies during training is key to better collaboration at test time. However, existing methods typically rely on trajectory-level proxies to measure diversity, which may label two distinct trajectories of the same strategy as diverse.
Instead, our key insight is to leverage LLMs' ability to describe behaviorally distinct agents in environments, similar to how humans can recognize high-level strategies. To this end, we introduce \popgen{}, a framework that synthesizes \textit{semantically} diverse agents from scratch using LLMs. \popgen{} first generates a diverse set of textual descriptions of agent behaviors, then translates each behavior into a code policy that can be executed within the environment.
We show that cooperators trained against \popgen{} populations achieve robust zero-shot cooperation with unseen partners, outperforming population-play baselines across four collaborative environments. We further show that semantic diversity scales with \popgen{} population size, improving the cooperator’s worst-case performance, and we identify design choices that preserve population quality while maximizing diversity.
Our code is available at \url{https://github.com/labicon/PopGen}.
\end{abstract}
\vspace{-2mm}
\section{Introduction}
\label{sec:introduction}

\suppressfloats[t]
\begin{figure}[t]
    \centering
    \includegraphics[width=\linewidth]{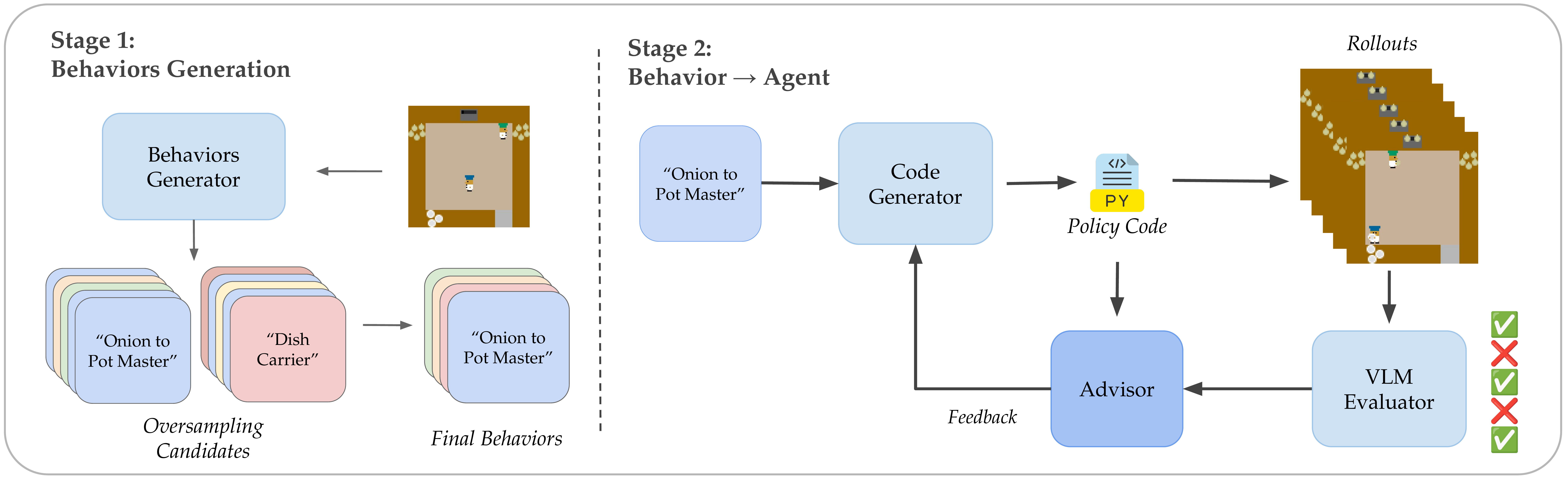}
    \caption{\textbf{Overview of \popgen{}.} \textit{Left (Stage~1):} An LLM produces a diverse set of behaviors at the text level using oversampling. \textit{Right (Stage~2):} Each behavior description enters an iterative refinement loop: a generator translates the behavior into an executable policy (Policy Code), which is evaluated with a VLM through environment rollouts. An advisor then analyzes rollout results and provides feedback to guide subsequent refinement attempts.}
    \label{fig:popgen}
    \vspace{-15pt}
\end{figure}

Training agents that collaborate effectively with unseen partners is a key challenge in multi-agent systems like robotics or cooperative video games. The task, known as zero-shot coordination (ZSC), requires a cooperator to coordinate with partners it has never interacted with, whose conventions, skill level, and strategy are unknown at test time. This problem is challenging because cooperative tasks typically admit many equally effective but mutually incompatible conventions, so a cooperator trained with one arbitrary convention may fail when paired with a partner that follows a different one~\citep{hu2020other, carroll2019utility}.

A common approach to ZSC is population play, which trains a cooperator against a population of partner agents. The effectiveness of population play depends on the \emph{diversity} of the population against which the cooperator is trained, since a high coverage of the strategy space during training increases the likelihood that the test-time partner is in-distribution for the cooperator, allowing them to collaborate effectively~\citep{strouse2021collaborating, lupu2021trajectory, sarkar2023diverse}. Existing methods build such populations by optimizing diversity proxies computed from trajectory statistics~\citep{lupu2021trajectory, zhao2022maximum}. However, these measures may fail to capture what humans perceive as meaningfully distinct strategies~\citep{bettini2024neuraldiversitymeasuringbehavioral, wang2025diversityhumanfeedback}. For example, two chefs may take different trajectories in a kitchen while following the same strategy (e.g., only carrying onions). The issue is not that these trajectories are wrongly classified as different (they are indeed different), but that optimizing for such diversity can lead to undesired training populations that are diverse in their motion but do not cover the distinct strategies that matter for coordination~\citep{sarkar2023diverse}.

In contrast to these measures, humans naturally describe trajectories at a higher level, recognizing groups of strategies. Our key idea is to use LLMs as a higher-level interface for specifying differences through natural language, as humans do. This allows us to represent behaviors in terms of their high-level \textit{semantics} rather than their low-level actions, providing a natural space for expressing diverse behaviors. We introduce \popgen{}, a framework that leverages LLMs to define semantically diverse behaviors in language space and translate them into executable policies that can interact with the environment. \popgen{} operates in two stages. First, it generates a diverse set of behavior descriptions in natural language, where high-level semantic differences can be expressed explicitly. We use tailored prompting methods to balance the quality of behavior descriptions with their semantic diversity, encouraging descriptions that are both meaningful and genuinely distinct. Second, \popgen{} grounds these high-level semantic descriptions into executable agent behaviors by synthesizing policy execution code. We generate the policy code in an iterative loop, where structured feedback from an advisor module guides progressive refinement.

We use the diverse population generated by \popgen{} for population-play training and evaluate its impact on cooperation with novel partners. We find that training with semantically diverse partners improves robust cooperation, demonstrating the value of diverse behaviors. We further show that semantic diversity grows with larger \popgen{} populations, and that \popgen{} increases diversity while controlling for the quality of the synthetic agents~\citep{pugh2016qualitydiversity}. Finally, we show that turning textual behavior descriptions into executable policy code, rather than LLM-generated reward functions, yields partners that more faithfully exhibit their specified behaviors and leads to more robust cooperators.

\noindent In summary, our contributions are:
\begin{itemize}[nosep]
    \item We propose \popgen{}, a framework that generates diverse agent populations from scratch by leveraging LLMs' understanding of semantic diversity.
    \item We show that training a cooperator against a \popgen{} population achieves the best performance across four collaborative environments when matched against other population-play baselines.
    \item We provide extensive ablations that aim to understand the link between semantic diversity and downstream cooperation, as well as insights on methods to turn textual descriptions into executable agents using LLMs.
\end{itemize}
\vspace{-2mm}
\section{Related Work}
\label{sec:related_work}

\paragraph{Generating diverse populations for ZSC.}
Zero-shot coordination requires a cooperator to work with teammates it never saw during training~\citep{hu2020other}. Population play achieves this by training an agent against a set of partners, and how well population play works depends on the behavioral diversity of the population against which an agent is trained~\citep{bettini2024impact, mckee2022quantifying}. Existing methods for generating agent populations vary from pooling checkpoints across random seeds~\citep{strouse2021collaborating} to optimizing explicit diversity objectives over trajectory or action distributions~\citep{lupu2021trajectory, zhao2022maximum}, and more recently to sampling partners from a generative model of human play~\citep{liang2024gamma}. These diversity measures all operate in trajectory, action, or latent space, where distances may not reflect the strategic distinctions a human would recognize~\citep{bettini2024neuraldiversitymeasuringbehavioral, wang2025diversityhumanfeedback}. We instead leverage the semantic diversity captured at the text level to generate diverse populations of agents.

\paragraph{LLM-generated agents.}
LLMs can translate natural-language specifications into agents in several forms. LLM-as-Agent systems directly interact with the environment but require costly LLM calls during rollout~\citep{huang2022language, yao2023react}, while reward-generation methods translate textual objectives into reward functions and train RL policies~\citep{ma2023eureka, xie2023text2reward}. However, designing rewards that faithfully capture high-level behaviors is challenging, as RL policies can optimize the reward without exhibiting the intended behavior~\citep{choi2025craft}. This challenge is pronounced in multi-agent settings, where interactions depend on other agents. Policy code instead provides a direct representation of behavior by explicitly encoding its decision logic~\citep{liang2023code, wang2023voyager, fu2026capx, ahn2026towards}. Recent advances in LLM code generation further enable refining such policies through execution feedback~\citep{fu2026capx, ahn2026towards}. In our experiments, we show that Policy Code more reliably realizes the specified behaviors than RL policies, resulting in higher behavioral adherence and stronger cooperator performance.

\vspace{-2mm}
\section{Problem Formulation}
\label{sec:problem_formulation}



\paragraph{Multi-agent interaction.}
To formalize multi-agent coordination, we adopt the Dec-POMDP framework~\citep{bernstein2002complexity, oliehoek2016concise}. A Dec-POMDP is a tuple $\mathcal{M} = \langle \mathcal{I}, \mathcal{S}, \{\mathcal{A}^i\}_{i \in \mathcal{I}}, \{\mathcal{O}^i\}_{i \in \mathcal{I}}, \{Z^i\}_{i \in \mathcal{I}}, T, R, \gamma \rangle$, where $\mathcal{I} = \{1, \dots, n\}$ is the set of agent indices and $\mathcal{S}$ is the state space. Each agent $i$ has an action space $\mathcal{A}^i$ and an observation space $\mathcal{O}^i$. From state $s_t$ at time step $t$, the observation function $Z^i$ defines the probability of agent $i$ receiving an observation $o_t^i \sim Z^i(\cdot \mid s_t)$. An action $a_t^i$ of agent $i$ is sampled from its policy $\pi^i(\cdot \mid h_t^i)$, where $h_t^i = (o_0^i, \ldots, o_t^i)$ is agent $i$'s observation history. The transition function $T(\cdot \mid s_t, a_t)$ defines the probability of the next state $s_{t+1}$ given the joint action $a_t = (a_t^1, \dots, a_t^n)$. $R$ is the shared reward function $r_t = R(s_t, a_t)$, and $\gamma \in [0,1)$ is the discount factor.

\paragraph{Training against populations of partners.}
In this work, we consider a symmetric two-player setting in which the two agents are exchangeable. For notational simplicity, we designate agent $1$ as the cooperator and agent $2$ as the partner. Population play proceeds in two phases. It first constructs a population $\Pi = \{\pi^{2}_1, \ldots, \pi^{2}_K\}$ of $K$ partner policies that exhibit diverse coordination behaviors. In the second phase, we train a cooperator policy $\pi^1_\phi$ with parameters $\phi$ to maximize the expected return averaged over partners sampled from $\Pi$:
\begin{equation}
    \phi^* = \arg\max_{\phi} \; \mathbb{E}_{\pi^{2}_k \sim \text{Uniform}(\Pi)}\left[J(\pi^1_\phi, \pi^{2}_k)\right] = \arg\max_{\phi} \; \frac{1}{K}\sum_{k=1}^{K} J(\pi^1_\phi, \pi^{2}_k),
    \label{eq:pp}
\end{equation}
where $J(\pi^1_\phi, \pi^{2}_k) = \mathbb{E}\big[\sum_{t=0}^{\infty} \gamma^t r_t\big]$ is the expected discounted shared return when the cooperator $\pi^1_\phi$ is paired with partner $\pi^{2}_k$. In \popgen{}, each partner policy $\pi^2_k$ is associated with a natural-language behavior description $b_k$. We denote the corresponding behavior set by $\mathcal B = \{b_1,\ldots,b_K\}$.

\section{\popgen{}: Generating Diverse Agent Populations}
\label{sec:popgen}

We introduce \popgen{}, a framework for synthesizing semantically diverse populations in multi-agent environments. The resulting population can then be used to train a cooperator for zero-shot cooperation. As illustrated in Figure~\ref{fig:popgen}, \popgen{} leverages LLMs in two stages. 


\subsection{Stage 1: Generating Semantically Diverse Behaviors}
\label{sec:method_behaviors}
The first stage feeds an environment description as input to an LLM, and queries it to generate various behaviors that agents could elicit in it (e.g., an ``onion-to-pot master'' in \environment{overcooked}). 
The objective here is to generate a pool of \textit{diverse} behaviors, in order to train a cooperator against it~\citep{lupu2021trajectory,bettini2024impact}. Such diversity is hard to measure directly from policies, since two policies can take different actions while following the same strategy~\citep{bettini2024neuraldiversitymeasuringbehavioral, wang2025diversityhumanfeedback}. We instead measure \textit{semantic} diversity from the behaviors' textual descriptions. Specifically, we encode each description $b_k$ into a unit-normalized text embedding $e_k$ and quantify the diversity of a behavior set $\mathcal{B} = \{b_1, \ldots, b_K\}$ as the average pairwise distance between its embeddings:
\begin{equation}
    D(\mathcal{B}) = \frac{2}{K(K-1)} \sum_{k < \ell} \sqrt{2\big(1 - e_k^\top e_\ell\big)}.
    \label{eq:diversity}
\end{equation}
\paragraph{The tension between quality and diversity of agent populations.} An important challenge when targeting textual diversity in the population is to avoid achieving it at the cost of degrading the quality of the generations. For instance, one can simply generate diverse but nonsense text as the behaviors for an environment to optimize diversity. This problem, known as the quality-diversity trade-off in NLP~\citep{zhang2020tradingdiversityqualitynatural}, is a result of the two metrics directly conflicting when optimizing for diversity\footnote{The intuition is the following: optimizing diversity means expanding towards uncovered axes \textit{unconditionally}, while optimizing quality \textit{restricts} the search space to fewer points.}. In our setup, we define behavior \textit{quality} along the following axes: \textit{realism} (would a real player plausibly play this way?); \textit{relevance} (is the behavior relevant to the environment?); and \textit{specificity} (is the behavior specific enough so that it encodes a recognizable style of play?). Concretely, we compute this by grading each axis from 0 to 5 with an LLM\footnote{The judge prompt is detailed in Appendix~\ref{app:quality_prompt}.}, and sum these into a score out of 15.

\popgen{} is carefully designed to address this tension, decoupling the search for diverse behaviors from the selection of relevant ones to preserve quality of generations. The LLM is first queried to generate a set of candidates by oversampling (i.e., generating many more candidates than expected), then re-queried to select the highest-quality candidates among the set as shown in Figure~\ref{fig:popgen}. Additionally, \popgen{} employs a simple trick from Verbalized Sampling (VS), which queries the LLM to output each candidate's probability alongside the generation and has shown strong empirical results~\citep{zhang2025verbalizedsampling}.
We empirically show that \popgen{} can generate behavior populations that are both high-quality and semantically diverse in Section~\ref{sec:diverse_behaviors}.

\subsection{Stage 2: Instantiating Behaviors as Agents}
\label{sec:method_agents}
The second stage translates each textual behavior into a policy that interacts with the environment and other agents using an LLM. An LLM can act as the policy itself, outputting an action at every step~\citep{yin2025context, wang2024prompt}, generate a reward function for an RL policy to optimize~\citep{ma2023eureka, xie2023text2reward, li2025semdiv}, or directly implement the decision logic as executable code~\citep{liang2023code, wang2023voyager}. \popgen{} adopts the latter, modeling each agent's policy as a Python function, which is cheap to execute and avoids training an RL agent per behavior. Section~\ref{sec:pc_v_rf} compares this choice with SemDiv's reward-function approach~\citep{li2025semdiv}.

\paragraph{Measuring adherence with an evaluator.}
A key requirement in Stage~2 is to evaluate whether a generated policy actually follows its target behavior. For this, we define \emph{adherence}. Adherence is measured by rolling out a candidate policy $\pi$ with a partner, and asking a \emph{VLM} to classify whether the trajectory $\tau$ aligns with the intended behavior $b$ based on screenshots and state sequences of $\tau$. The classifier outputs $1$ if a trajectory aligns with $b$ and $0$ otherwise. We repeat this on $M$ different initial environment states to prevent a candidate from achieving high adherence only under a favorable configuration~\citep{zheng2023judging, choi2025craft}, and report the final score as the fraction of accepted rollouts $\tau_1, \dots, \tau_M$:

\begin{equation}
    \mathrm{Adh}(\pi, b) = \frac{1}{M} \sum_{m=1}^{M} \text{VLM}(\tau_m, b) \in [0, 1].
    \label{eq:adherence}
\end{equation}



\paragraph{Iterative refinement loop.}

Generating a policy that reliably exhibits a specified behavior can be challenging in complex multi-agent environments~\citep{choi2025craft}. We address this challenge with an iterative refinement procedure~\citep{madaan2023selfrefine, khattab2024dspy} based on the above adherence measure. The LLM iteratively generates a candidate policy from the behavior and environment description, and gets rich feedback from a separate LLM advisor based on candidate policy rollouts. This loop is repeated until the candidate policy reaches a target level of adherence or all attempts are exhausted. The refinement loop is detailed in Algorithm~\ref{algo:refinement_loop}.

\subsection{Population Play with \popgen{} Partners}
\label{sec:method_cooperator}
A \popgen{} population can serve directly as the partner population for population play. We call this \popplay{}, which trains a cooperator $\pi^1_\phi$ against a \popgen{} partner population. 
Given partner policies $\Pi_{\text{\popgen{}}} = \{\pi^{2}_1, \ldots, \pi^{2}_K\}$, we optimize the cooperator's expected return over sampled partners with PPO~\citep{schulman2017ppo}. The partner policies are frozen, so only $\pi^1_\phi$ is updated. Because \popgen{} partners are executable Policy Code, they add negligible overhead to the training loop compared to sampling from a learned generative model or maintaining a live population of RL agents. Rather than sampling partners uniformly during training, we draw them using the rank-based prioritized scheme of MEP~\citep{zhao2022maximum}, which samples partners that the cooperator performs poorly with at higher probability, preventing the cooperator from overfitting to easy-to-coordinate partners while neglecting hard ones (Appendix~\ref{app:baselines}). 

\vspace{-2mm}
\section{Experiments}
\label{sec:experiments}

\begin{figure}[t]
    \centering
    \includegraphics[width=\textwidth]{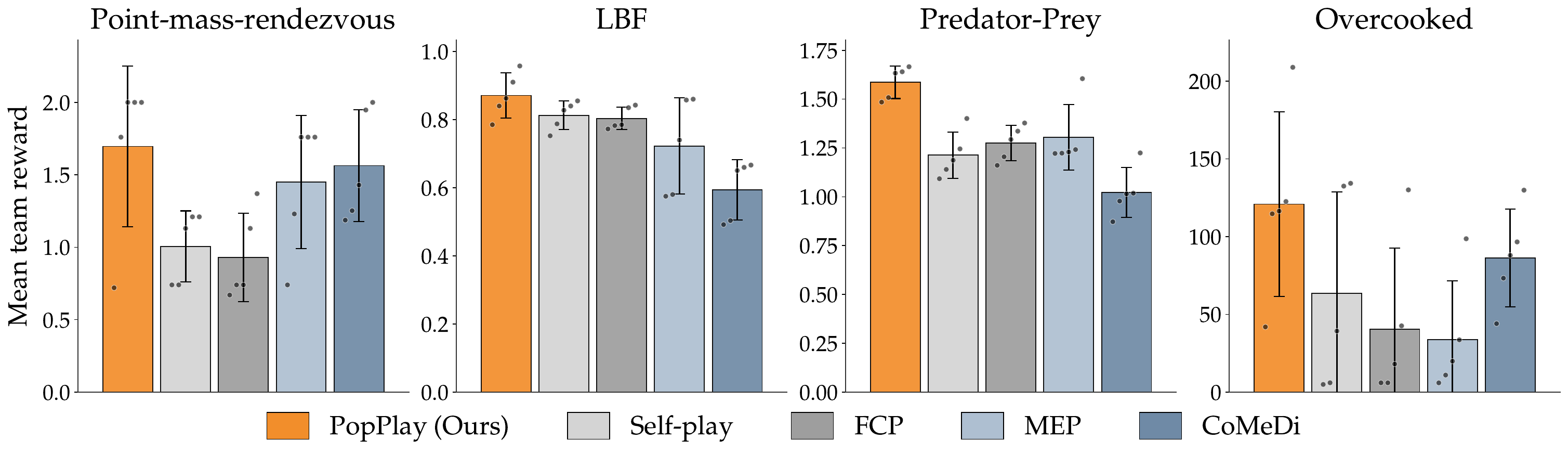}
    \caption{\textbf{Zero-shot coordination with held-out partners.} Each bar reports the average team reward (higher is better) when each cooperator is paired with the held-out partners, which do not appear in any training population. Error bars give the standard deviation across five runs, and dots mark individual runs. \popplay{} attains the highest reward in all four environments. Table~\ref{tab:coop_results_heldout_partners} reports the reward against each individual partner.}
    \label{fig:zs_cooperation}
    \vspace{-4mm}
\end{figure}

\begin{figure}[t]
    \centering
    \begin{minipage}[t]{0.45\linewidth}
        \vspace{0pt}
        \centering
        \subfloat[Rendered rollouts of \popplay{} cooperator (red box) paired with a partner. \popplay{} switches roles based on the partner's behavior.\label{fig:oc_rollout}]{%
            \includegraphics[width=\linewidth]{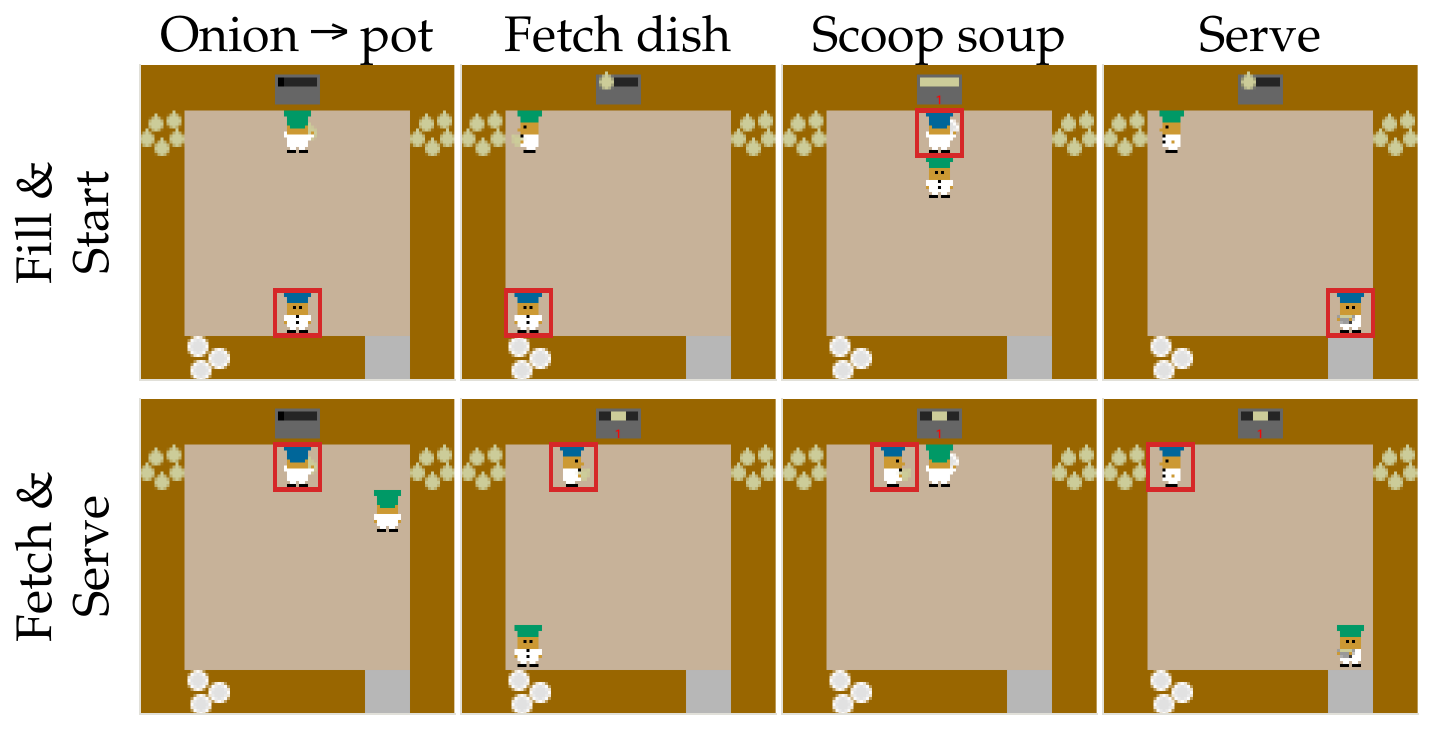}}
    \end{minipage}
    \hfill
    \begin{minipage}[t]{0.53\linewidth}
        \vspace{0pt}
        \centering
        \subfloat[Share of each subtask completed by the cooperator, averaged over the four
                  held-out partners, with the subtask count per episode in parentheses.%
                  \label{tab:oc_events}]{%
        \footnotesize
        \setlength{\tabcolsep}{2.5pt}
        \begin{tabular}{lcccc}
        \toprule
        \textbf{Method} & \makecell{Onion\\$\rightarrow$ pot} & \makecell{Start\\pot} & \makecell{Scoop\\soup} & \makecell{Serve\\soup} \\
        \midrule
        \popplay{} & $26\%$ {\scriptsize$(2.7)$}  & $26\%$ {\scriptsize$(2.7)$} & $27\%$ {\scriptsize$(2.8)$} & $21\%$ {\scriptsize$(2.2)$} \\
        \midrule
        Self-Play  & $0\%$ {\scriptsize$(0.0)$}   & $0\%$ {\scriptsize$(0.0)$}  & $56\%$ {\scriptsize$(1.8)$} & $44\%$ {\scriptsize$(1.4)$} \\
        FCP        & $0\%$ {\scriptsize$(0.0)$}   & $0\%$ {\scriptsize$(0.0)$}  & $62\%$ {\scriptsize$(1.2)$} & $38\%$ {\scriptsize$(0.7)$} \\
        MEP        & $9\%$ {\scriptsize$(0.2)$}   & $13\%$ {\scriptsize$(0.3)$} & $52\%$ {\scriptsize$(1.1)$} & $26\%$ {\scriptsize$(0.5)$} \\
        CoMeDi     & $68\%$ {\scriptsize$(6.1)$}  & $21\%$ {\scriptsize$(1.9)$} & $6\%$ {\scriptsize$(0.5)$}  & $5\%$ {\scriptsize$(0.5)$} \\
        \bottomrule
        \end{tabular}}
    \end{minipage}

    \vspace{0pt}
    \begin{minipage}[t]{\linewidth}
        \centering
        \subfloat[Worst-case team reward against held-out partners. For each run, we report the reward with the held-out partner on which the cooperator scores lowest (mean $\pm$ std over five runs).%
                  \label{tab:worst_partner}]{%
        \footnotesize
        \begin{tabular}{lcccc}
        \toprule
        \textbf{Method} & PMR & LBF & PP & Overcooked \\
        \midrule
        \popplay{} (ours) & $\mathbf{1.41 \pm 0.89}$ & $\mathbf{0.76 \pm 0.16}$ & $\mathbf{1.33 \pm 0.13}$ & $\mathbf{31.2 \pm 44.6}$ \\
        \midrule
        Self-Play (SP)    & $0.00 \pm 0.00$ & $0.71 \pm 0.11$ & $1.04 \pm 0.14$ & $0.0 \pm 0.0$ \\
        FCP               & $0.00 \pm 0.00$ & $0.73 \pm 0.08$ & $1.07 \pm 0.14$ & $0.0 \pm 0.0$ \\
        MEP               & $0.62 \pm 0.57$ & $0.68 \pm 0.14$ & $0.98 \pm 0.09$ & $2.3 \pm 3.7$ \\
        CoMeDi            & $1.40 \pm 0.52$ & $0.49 \pm 0.08$ & $0.64 \pm 0.21$ & $11.2 \pm 0.5$ \\
        \bottomrule
        \end{tabular}}
    \end{minipage}

    \vspace{-4pt}
    \caption{\textbf{Cooperator role adaptation in \environment{overcooked} and robustness
    to held-out partners.} (a) Rollouts show \popplay{} taking semantically diverse behaviors interpretable to humans.
    (b) \popplay{} spreads its actions across subtasks while the baselines tend to focus on a subset of actions. Per-partner counts are in Table~\ref{tab:oc_events_full}. (c) \popplay{} achieves the highest worst-case reward in all environments.}
    \label{fig:overcooked_roles}
    \vspace{-15pt}
\end{figure}

We evaluate \popgen{} as a method for generating synthetic populations to train cooperators. After describing the setup (Section~\ref{sec:exp_setup}), we test whether cooperators trained against \popgen{} populations generalize to novel partners (Section~\ref{sec:coop_results}), analyze how \popgen{} produces semantically diverse agents and why this helps cooperation (Section~\ref{sec:diverse_behaviors}), and compare policy code against LLM-generated reward functions for implementing behaviors (Section~\ref{sec:pc_v_rf}).

\vspace{-2mm}
\subsection{Experimental Setup}
\label{sec:exp_setup}

\paragraph{Environments.}
We consider four multi-agent environments with discrete and continuous action spaces. In Point-Mass Rendezvous (\environment{PMR})~\citep{lowe2017multi}, two agents must agree on which of four equally valid landmarks to meet at. In Level-Based Foraging (\environment{LBF})~\citep{albrecht2015gametheoreticmodelbestresponselearning}, agents must time their load actions to jointly collect food whose level exceeds any single agent's. In Predator-Prey (\environment{PP})~\citep{lowe2017multi}, agents choose between capturing low-reward rabbits alone and coordinating to capture high-reward stags. In \environment{overcooked}~\citep{carroll2019utility}, two chefs must divide subtasks (filling pots with onions, cooking, serving) to deliver soups. Details are in Appendix~\ref{app:popgen_envs}.

\paragraph{LLM models.} We employ Gemini-3-Flash as the generator model and adherence classifier, and Gemini-3-Pro as the feedback advisor (see Figure~\ref{fig:popgen}) in our experiments~\citep{google2025gemini3pro}. We use Gemini-Embedding-001 to obtain behavior embeddings, which we use to measure semantic diversity.


\paragraph{Cooperator training.}
We compare \popplay{}, which trains a cooperator via population play against \popgen{} partners, with four population-play baselines. \textit{Self-Play (SP)}~\citep{strouse2021collaborating} trains MAPPO self-play agents from independent seeds. \textit{Fictitious Co-Play (FCP)}~\citep{strouse2021collaborating} adds their intermediate checkpoints to cover a range of skill levels. \textit{Maximum Entropy Population-based Training (MEP)}~\citep{zhao2022maximum} diversifies the population with a population-entropy objective and trains the cooperator with prioritized partner sampling. \textit{CoMeDi}~\citep{sarkar2023diverse} adds partners that minimize cross-play reward with the existing population, encouraging incompatible conventions, and trains the cooperator with an auxiliary behavior-cloning loss on each partner's self-play rollouts. All cooperators except CoMeDi share the same PPO~\citep{schulman2017ppo} backbone and number of training frames. Appendices~\ref{app:baselines} and~\ref{app:cooperator} detail the baselines and their hyperparameters in our setup.

\paragraph{ZSC evaluation.} We evaluate each cooperator's zero-shot coordination by pairing it with \emph{held-out partners} that were not encountered during training. These are fixed, scripted agents with semantically distinct behaviors, specified in advance without LLMs. The held-out behaviors are never provided to the LLMs during our experiments, preventing \popgen{} from being biased toward the evaluation behaviors. Appendix~\ref{app:eval_partners} lists the held-out partners for each environment.

\vspace{-2mm}
\subsection{Training Cooperators with \popgen{} Populations}
\label{sec:coop_results}

We train a cooperator against a \popgen{} population (referred to as \popplay{}), and compare how well it coordinates with held-out partners against the baselines. We then analyze the resulting trajectories in \environment{overcooked} to see how each cooperator divides the work with its partner.


\paragraph{Generalization to novel partners.}
Figure~\ref{fig:zs_cooperation} reports the average team reward across different held-out partners of each method (mean and standard deviation over five runs). \popplay{} achieves the highest mean return in every environment, clearly outperforming the baselines in \environment{LBF} and \environment{PP}, while attaining a higher mean return than the strongest baseline, CoMeDi, in \environment{PMR} and \environment{overcooked}, with overlapping error bars. This consistently high mean performance indicates robust generalization across held-out partners. To further assess robustness to individual partners, Figure~\ref{tab:worst_partner} reports the team reward when each cooperator is paired with its worst-performing held-out partner. \popplay{} maintains the highest worst-case rewards, whereas Self-Play and FCP fall to zero reward with their worst partners in \environment{PMR} and \environment{overcooked}. Table~\ref{tab:coop_results_heldout_partners} reports the breakdown of average team reward for each partner. 

\paragraph{\popplay{} adapts its role to the partner.} We examine whether \popplay{} can show semantically different behaviors when paired with different partners. \environment{overcooked} provides a natural setting for this analysis, as it requires a cooperator to handle different subtasks according to the partner chef's preference. Figure~\ref{fig:oc_rollout} illustrates this behavior by showing the \textit{same} \popplay{} cooperator paired with different partners. When the partner fetches onions, \popplay{} takes the dish and serves the soup. When the partner takes the dish, \popplay{} fetches onions and cooks instead. Figure~\ref{tab:oc_events} shows the distribution of subtasks completed by the cooperator. The \popplay{} cooperator performs all subtasks with substantial count, while other methods almost never load the onion (Self-Play, FCP, and MEP) or are biased toward single actions (CoMeDi on pot loading). In particular, Self-Play and FCP exhibit a severe mode collapse in population generation where their generated populations predominantly focus on onion fetching. As a result, the corresponding cooperators rarely need to fetch onions themselves, leading to the absence of onion-fetching behavior with held-out partners. We study these failures in detail in Appendix~\ref{app:sp_collapse}, highlighting the importance of behavioral diversity in the training population for learning robust partner-adaptive behavior.

\begin{figure}[t]
    \centering
    \begin{minipage}[t]{0.435\linewidth}
        \centering
        \subfloat[Diversity of the training pool and worst-case team reward as a function of population size.\label{fig:population_scaling}]{%
            \includegraphics[width=\linewidth]{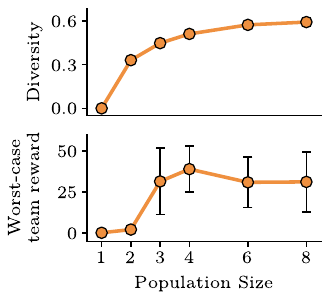}}
    \end{minipage}
    \hfill
    \begin{minipage}[t]{0.525\linewidth}
        \centering
        \subfloat[Team reward with held-out partners as a function of training pool diversity.\label{fig:diversity_reward}]{%
            \includegraphics[width=\linewidth]{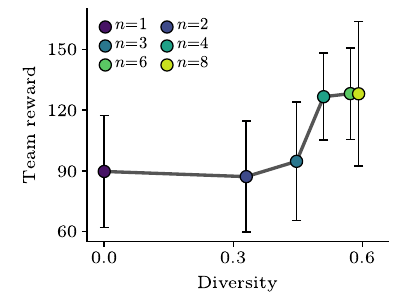}}
    \end{minipage}

    \vspace{-4pt}
    \caption{\textbf{Scaling the \popgen{} population in \environment{overcooked}.} \popplay{} cooperators trained against pools of $n$ distinct \popgen{} agents at a fixed budget. (a) Larger populations are more diverse and raise the worst-case team reward with held-out partners. (b) Higher diversity also raises the average team reward, with gains plateauing beyond $n=4$. Error bars show the standard error over runs.}
    \label{fig:training_diversity}
    \vspace{-15pt}
\end{figure}

\vspace{-2mm}
\subsection{Generating Diverse Populations for Cooperator Training}
\label{sec:diverse_behaviors}


Diversity in training agents is the core element driving cooperator performance as it allows the cooperator to adapt to a wide range of partners at test time~\citep{lupu2021trajectory, bettini2024impact}. In this section, we evaluate the diversity of \popgen{} populations. First, we investigate how diversity scales with \popgen{} population size, and measure its downstream impact on the cooperator's ability to adapt to new partners. We then evaluate how our generation method targets high semantic diversity while maintaining the quality of the behaviors (i.e., the quality-diversity trade-off introduced in Section~\ref{sec:method_behaviors}).

\paragraph{Larger populations for more diversity, and more robustness.}
We train \environment{overcooked} cooperators against \popgen{} populations with $n \in \{1, 2, 3, 4, 6, 8\}$ distinct behaviors. The training budget is fixed for fair comparison, so every pool contains 8 partners obtained by replicating the $n$ distinct agents (i.e., $n = 1$ corresponds to 8 copies of a single partner). For each size, we run the experiment 5 times (i.e., we generate 5 independent populations and train one cooperator on each), so the variation we report covers the population generation.

As shown in Figure~\ref{fig:population_scaling}, semantic diversity increases with larger \popgen{} population sizes. It grows from 0 at $n = 1$ to 0.51 at $n = 4$, and saturates afterwards as the first behaviors already cover most of the environment. These gains in diversity naturally translate into better cooperator performance with new partners~\citep{bettini2024impact}, with an inflection point suggesting that the population captures the diversity needed to represent most aspects of \environment{overcooked} (Figure~\ref{fig:diversity_reward}, $+34\%$ performance when adding the 4th agent). More importantly, larger populations make the cooperator more robust. We measure it as the reward with the held-out partner each cooperator coordinates with least (Figure~\ref{fig:population_scaling}, bottom). With $n \leq 2$, every cooperator fails completely with at least one held-out partner (worst-case reward around 0 in all runs), whereas from $n = 3$ onwards, worst-case reward significantly jumps (up to 39 on average). This last result means that a larger population is more likely to include a partner resembling any given unseen one, so the cooperator is less likely to face a partner it has never learned to complement.

\begin{wrapfigure}{r}{0.45\textwidth}
    \vspace{-\intextsep}
    \centering
    \includegraphics[width=\linewidth]{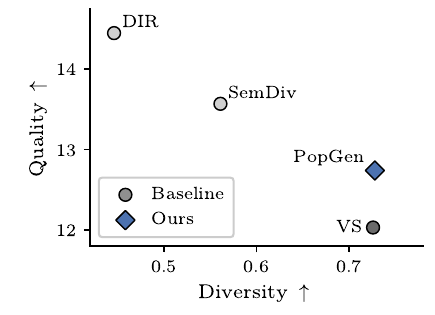}
    \caption{\textbf{Diversity and quality of \popgen{} population agents.} Quality and diversity of a generated set of 30 behaviors, averaged across the environments. \popgen{} matches the diversity of VS while improving quality, and is substantially more diverse than DIR and SemDiv.}
    \label{fig:behavior_generation_pareto}
    \vspace{-\intextsep}
\end{wrapfigure}
\paragraph{Preserving training agents' quality while increasing diversity.} Naively pushing for diversity tends to erode quality, as introduced in Section~\ref{sec:method_behaviors}. However, preserving quality matters for cooperator training, as incoherent or degenerate behaviors provide a noisy signal and can lead the cooperator to learn compensatory strategies that are useless, or even harmful, when paired with realistic partners~\citep{liang2024gamma}. We evaluate \popgen{}'s ability to preserve a correct level of quality in agent behaviors while aiming for semantic diversity. 
We compare \popgen{} against three baselines. \textit{DIR} generates each behavior independently, with no knowledge of previous generations, and thus cannot reach high diversity but serves as a quality oracle. \textit{Verbalized Sampling (VS)}~\citep{zhang2025verbalizedsampling} is a method designed to preserve quality when generating diverse outputs, and was evaluated on standard NLP tasks (e.g., QA, story generation). Finally, \textit{SemDiv}~\citep{li2025semdiv} is the pipeline closest to ours: it generates textual behaviors sequentially, conditioning each new behavior on all previous ones. For each method, we generate a set of 30 behaviors for each environment listed in Section~\ref{sec:exp_setup}, and measure \textit{diversity} and \textit{quality}\footnote{We employ Gemini-3-Pro as the judge for quality.} as defined in Section~\ref{sec:method_behaviors}.

Figure~\ref{fig:behavior_generation_pareto} presents the results. \popgen{} achieves the best trade-off overall. It reaches the highest diversity ($0.728$), on par with VS ($0.726$), while producing clearly better behaviors at that diversity level ($12.74$ vs.\ $12.03$, a $6\%$ improvement in quality). Compared to SemDiv, \popgen{} gives up $6\%$ of quality for a $30\%$ gain in diversity, which is considerable for the task. As expected, DIR obtains the highest quality ($14.45$) but by far the lowest diversity ($0.446$), with \popgen{} being $63\%$ more diverse.

\subsection{Modeling Agent Behaviors as Policy Code for Population Play}
\label{sec:pc_v_rf}
\begin{figure}[t]
    \centering
    \includegraphics[width=\linewidth]{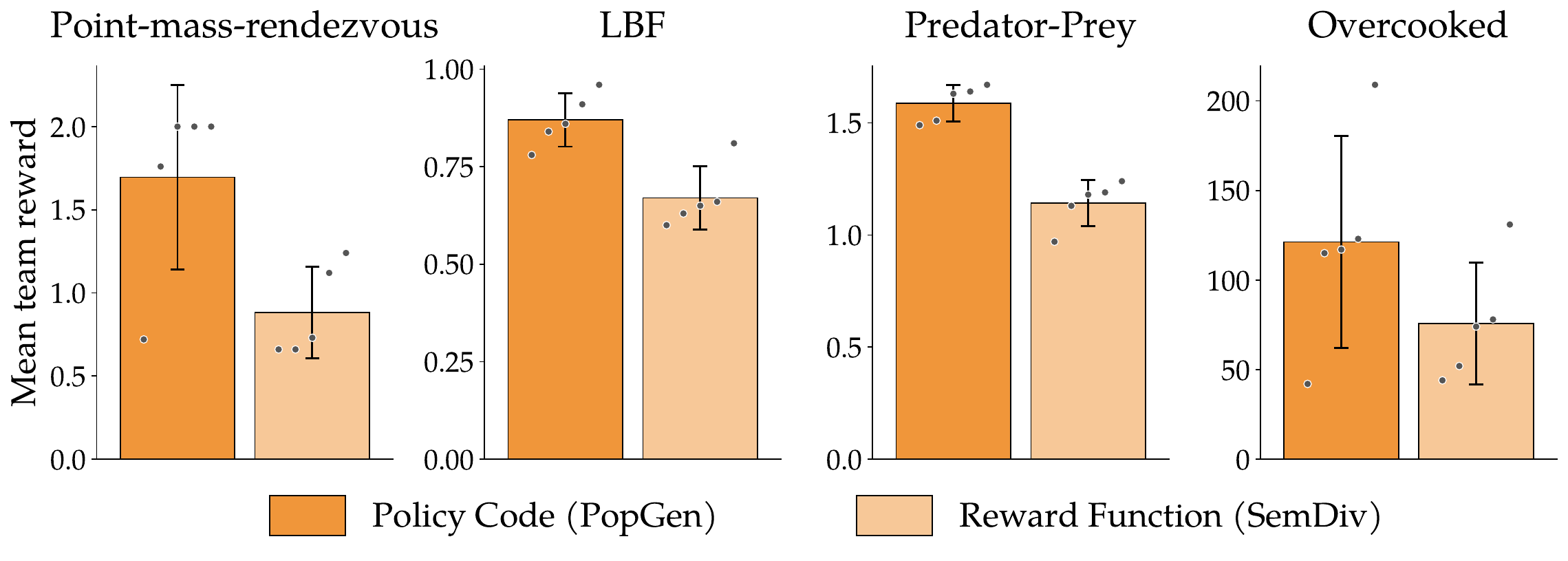}
    \caption{\textbf{Population play with policy code and reward-function partners.} Mean team reward against held-out partners of cooperators whose training partners are modeled as executable code (like \popgen{}) or as neural network policies trained with RL on LLM-generated reward functions (like SemDiv). The set of behaviors is kept identical, so that only agent modeling method changes. Error bars give the standard deviation across five runs, and dots mark individual runs.}
    \label{fig:pc_vs_rf}
    \vspace{-4mm}
\end{figure}

In its second stage, \popgen{} uses an LLM to turn each textual behavior description into an executable \textit{policy} that can act in the environment (see Figure~\ref{fig:popgen}).
In this section, we ablate this part, comparing \popgen{}'s approach, in which the LLM writes code that directly scripts the agent's actions (\emph{Policy Code}), and that of SemDiv~\citep{li2025semdiv}, in which the LLM writes a reward function used to train a neural agent to exhibit the behavior (\emph{Reward Function}). Both methods start from the same set of textual behaviors, which removes stage~1 as a potential confounder. We evaluate each method on two metrics. The first is downstream performance: the team reward a cooperator trained with each population achieves when paired with new partners. The second is the population's average adherence to its target behaviors (Section~\ref{sec:method_agents}), which tells us how faithfully each method models the behaviors in the policies, and lets us relate adherence to cooperation.

Figure~\ref{fig:pc_vs_rf} shows that cooperators trained with Policy Code partners (as in \popgen{}) achieve higher test-time reward in all four environments. This mirrors the partners' adherence: Policy Code partners adhere better on average ($0.66 \pm 0.07$ vs.\ $0.43 \pm 0.07$; Table~\ref{tab:b2t_all}). The adherence gap is widest in the discrete environments, \environment{LBF} ($0.61$ vs.\ $0.13$) and \environment{Overcooked} ($0.63$ vs.\ $0.41$), which are also where the Reward Function cooperator struggles most. Policy Code is also more time-efficient at aligning partners with a target behavior (Appendix~\ref{app:b2a_methods}).

\vspace{-3mm}
\section{Conclusion}
\label{sec:conclusion}

In this work, we study \emph{semantic} diversity, rather than trajectory-level statistics, as a basis for training agents that cooperate robustly with unseen partners. We introduce \popgen{}, which constructs diverse populations by first generating diverse behaviors in text space and then translating them into executable agents through advisor-guided refinement. We show that cooperators trained against \popgen{} populations generalize to new partners, outperforming population-play baselines in all four environments studied. We further study the link between semantic diversity and zero-shot cooperation, and show that training against semantically diverse partners improves the cooperator's worst-case performance by exposing it to a wide range of strategies. We also provide insights on how to preserve the quality of the generated agents (e.g., keeping behaviors realistic) while optimizing for diversity. Finally, we show that implementing behaviors as executable policy code, rather than as LLM-generated reward functions, yields partners that follow their intended behaviors more faithfully, particularly in discrete environments, and leads to more robust cooperators.

\paragraph{Limitations.}
Our work focuses on increasing \emph{behavior coverage}, but we evaluate only against synthetic partners. Prior work reports that training against human-like populations transfers to coordination with humans~\citep{carroll2019utility, liang2024gamma}, but a human study remains necessary to confirm that the semantic diversity of \popgen{} also improves human-AI coordination. In addition, because \popgen{} relies on LLMs for both behavior and policy generation, the generated populations, and hence cooperator performance, may vary across runs.

\paragraph{Future work.}
Two directions follow naturally. First, \emph{online adaptive behavior generation} would move behavior generation inside the cooperator training loop, so that \popgen{} can target the cooperator's current weaknesses~\citep{zhao2022maximum}. Second, the quality of the refinement loop is bounded by the context provided to the LLM. Advances in context engineering~\citep{mei2025contextengineering, khattab2024dspy, agrawal2026gepareflectivepromptevolution} and in code LLMs and coding agents~\citep{chen2021codex, roziere2024codellamaopenfoundation, kimiteam2026kimik3openfrontier} offer a direct path to improving both the advisor's feedback and the generator's code.



\section*{Acknowledgments}
This work benefited from valuable feedback from Antoine Bosselut and insightful discussions with Seoyeon Choi. We also thank Google for supporting this project.

\section*{AI Use Statement}

In this work, we used generative AI tools to propose or refine hypotheses, design or provide feedback on research methodology or experiments, implement methods, support qualitative and thematic data analysis, and interpret results. We have not used generative AI tools to help develop theoretical models or conceptual frameworks, formulate mathematical claims, or provide critical ingredients for proving mathematical claims, and generating synthetic data sets, assisting in the writing of proofs, assisting with translation, and cleaning and reformatting datasets are not applicable to this work. Additionally, we used generative AI tools to create or modify scientific figures or images, suggest experimental parameters, create or edit software code, brainstorm, edit a research paper to improve readability, and format references. We have reviewed all AI-assisted work. LLM-generated code and figures were verified by the authors, and the authors manually checked all AI-assisted paper writing. We take responsibility for the final content of this work, including text, claims or artifacts produced with the aid of generative AI.

\section*{Reproducibility Statement}
The behavior generation and refinement pipeline is specified in Section~\ref{sec:popgen}, with prompts detailed in Appendix~\ref{app:prompts}. Environments are described in Appendix~\ref{app:popgen_envs}, and held-out evaluation partners, cooperator baselines, and hyperparameters in Appendix~\ref{app:cooperator_training}. The per-environment numbers behind Figure~\ref{fig:zs_cooperation} are tabulated in Appendix~\ref{app:cooperator_results}. Source code is provided as supplementary material.

\bibliography{iclr2027_conference}
\bibliographystyle{iclr2027_conference}

\clearpage
\appendix
\begin{center}
    {\Large\sc Appendix}
\end{center}
\setcounter{topnumber}{4}
\setcounter{bottomnumber}{3}
\setcounter{totalnumber}{6}
\renewcommand{\topfraction}{0.95}
\renewcommand{\bottomfraction}{0.9}
\renewcommand{\textfraction}{0.05}
\renewcommand{\floatpagefraction}{0.75}

\section{\popgen{} Details}
\subsection{Environments}
\label{app:popgen_envs}
\begin{figure}[h]
    \centering
    \includegraphics[width=0.8\linewidth]{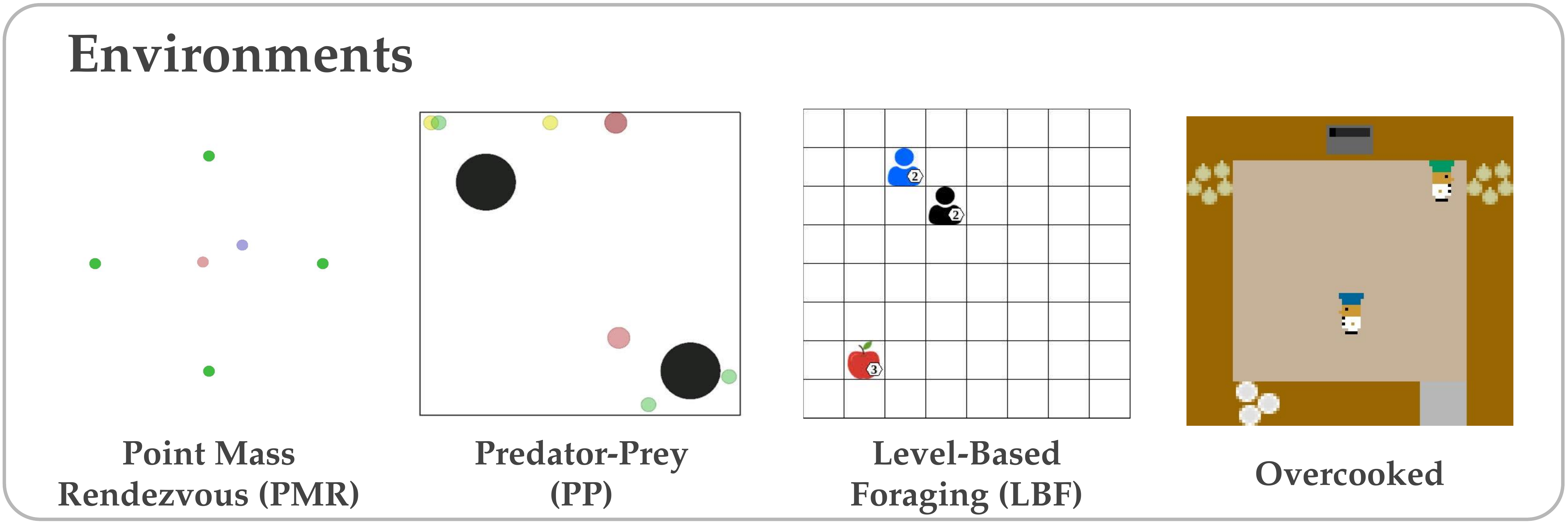}
    \caption{\textbf{Environments used for cooperator training.} From left to right, \environment{PMR} requires agents (red and purple) to navigate toward and converge on the same landmark (green). \environment{PP} requires agents (red) to coordinate to capture prey, with different rewards between individually capturing a rabbit (green) and jointly capturing a stag (yellow). \environment{LBF} requires agents to combine their carrying capacities to collect food items whose levels may exceed the level of a single agent. Finally, \environment{overcooked} requires two chefs to coordinate the sequential subtasks of preparing, cooking, and delivering soups.}
    \label{fig:environments}
\end{figure}

\begin{table}[h]
    \centering
    \caption{Multi-agent environments in \popgen{} cover cooperative settings across discrete and continuous action spaces. We note that adversarial agents (prey) in PP follow fixed policies defined by the environment, so only the ego team is generated and trained.}
    \label{tab:envs}
    \renewcommand{\arraystretch}{1.4}
    \resizebox{\linewidth}{!}{%
    \begin{tabular}{lllll}
    \toprule
    \textbf{Env} & \textbf{Num. Agents} & \textbf{Action Space} & \textbf{Observation} & \textbf{Objective} \\
    \midrule
    LBF      & 2 (ego: 2)            & Discrete (6)    & Partial & Collect all food items \\
    Overcooked & 2 (ego: 2)          & Discrete (6)    & Global  & Serve as many soups as possible \\
    PMR      & 2 (ego: 2)            & Continuous (2D) & Global  & Meet at the same landmark \\
    PP       & 7 (ego: 2, adv.: 5)   & Continuous (2D) & Partial & Capture prey \\
    \bottomrule
    \end{tabular}%
    }
\end{table}

Figure~\ref{fig:environments} and Table~\ref{tab:envs} show the environments used for cooperator training, covering different action spaces, agent compositions, observability, and reward structures.

\paragraph{Level-Based Foraging (LBF)~\citep{albrecht2015gametheoreticmodelbestresponselearning}.} \environment{LBF} is a cooperative gridworld in which agents navigate a discrete map and collect food items of varying levels. A team receives the collected food's level divided by the total food level in the field as a reward when they collect the food.

\paragraph{Overcooked~\citep{carroll2019utility}.} \environment{overcooked} places two chefs in a shared kitchen where they must fetch onions, load a pot, cook soup, plate it, and deliver it to the serving counter. They receive a reward for completing each subtask. A single delivery therefore chains several dependent subtasks, and either chef can perform any of them. This makes the division of labor, rather than individual skill, the main driver of team reward. We use the \emph{large\_room} layout in our experiments.

\paragraph{Point-Mass Rendezvous (PMR)~\citep{lowe2017multi}.} \environment{PMR} asks two point-mass agents to meet at the same landmark, with four landmarks placed evenly on a circle. The team receives a reward of $2$ only when both agents converge on the same landmark, and every landmark is equally valid.

\paragraph{Predator-Prey (PP)~\citep{lowe2017multi}.} \environment{PP} places agents in a continuous environment alongside multiple prey types with different capture requirements. Rabbits can be caught individually and provide a reward of $+0.5$, while stags require coordinated capture by multiple predators but yield a reward of $+2$. We train a team of $2$ predators while there are $3$ rabbits and $2$ stags in the field.

\subsection{Iterative Refinement for Adherence}
\label{app:popgen_evaluator}
\label{app:scripted_partner}

\paragraph{Evaluating adherence to behavior.} We estimate the adherence score in Equation~\ref{eq:adherence} with a VLM-as-a-judge that observes the trajectory and sampled frames from the rendered rollout and assigns a binary adherence label. The prompts passed to the VLM are available in Appendix~\ref{app:prompts}. When we roll out a candidate policy to measure adherence, we pair the generated agent with a fixed scripted partner. In \environment{LBF}, the partner moves to the closest food item and loads it. In \environment{PP}, it follows the scripted predator policy of~\citet{lowe2017multi}, which avoids solo attempts and moves in to coordinate a capture when a stag is present. In \environment{overcooked}, we use the oracle planning policy provided in the \texttt{overcooked\_ai} repository~\citep{carroll2019utility}. In \environment{PMR}, it selects a landmark uniformly at random and moves toward it. 

\paragraph{Feedback loop.} Algorithm~\ref{algo:refinement_loop} details the refinement loop employed in \popgen{}. Given a target behavior $b$ and environment description $s$, the \textsc{Generator} proposes a candidate policy $\pi_i$, which is executed for $M$ rollouts. A \textsc{VLM} judge inspects each rollout, deciding whether it exhibits $b$ and giving a rationale; the adherence score $\mathrm{Adh}_i$ is the fraction of rollouts judged consistent with $b$. If $\mathrm{Adh}_i \geq \alpha$, the policy is accepted immediately. Otherwise, an \textsc{Advisor} turns the judge's rationales, together with $b$, $s$, and the current policy, into targeted feedback $f_{i+1}$, and the generator regenerates the policy conditioned on that feedback. The loop runs for at most $I_{\max}$ iterations; if no attempt reaches the threshold, the highest-adherence policy is returned.

\begin{algorithm}[t]
\caption{Iterative Behavior-to-Agent Refinement}
\label{algo:refinement_loop}
\begin{algorithmic}[1]
\Require target behavior $b$, environment description $s$, adherence threshold $\alpha$, max iteration $I_{max}$, rollouts per attempt $M$
\Ensure best candidate policy $\pi^\star$ for behavior $b$
\State $f_1 \gets \emptyset$ 
\For{$i = 1, \dots, I_{max}$}
    \State $\pi_i \gets \textsc{Generator}(b, s, f_i)$ \Comment{regenerate given prior feedback}
    \State $\{\tau_m\}_{m=1}^{M} \gets \textsc{Rollout}(\pi_i, M)$
    \State $\mathrm{Adh}_i, \{\mathrm{rationale}_m\}_{m=1}^{M} \gets \textsc{VLM}(b, \{\tau_m\}_{m=1}^{M})$ \Comment{fraction of rollouts like $b$}
    \If{$\mathrm{Adh}_i \geq \alpha$}
        \State \Return $\pi_i$ 
    \EndIf
    \State $f_{i+1} \gets \textsc{Advisor}(b, s, \pi_i, \{\mathrm{rationale}_m\}_{m=1}^{M})$ \Comment{feedback for next attempt}
\EndFor
\State \Return $\pi_{i^\star}$ where $i^\star = \arg\max_{i \in \{1, \dots, I_{max}\}} \mathrm{Adh}_i$ \Comment{return best attempt}
\end{algorithmic}
\end{algorithm}

\paragraph{Advisor.} The advisor analyzes rollout evaluation results and produces structured feedback that guides the next refinement attempt. The prompts are shown in Appendix~\ref{app:prompts}.

\section{Cooperator Training}
\label{app:cooperator_training}

\subsection{Baselines}
\label{app:baselines}

Every baseline follows the two-stage population play recipe. The first stage builds a partner population $\Pi = \{\pi_1, \dots, \pi_K\}$, and the second stage trains a cooperator $\pi_\phi$ against it. The baselines differ mainly in how the first stage produces the partners, which is the stage \popgen{} replaces with language-guided policy code generation. Throughout this section we write $J(\pi, \pi')$ for the expected discounted team return of the pair $(\pi, \pi')$, so $J(\pi, \pi)$ is a \emph{self-play} return.

\paragraph{Self-Play (SP).} Self-Play trains one policy by pairing it with copies of itself, so run $k$ starting from seed $\xi_k$ solves $\pi_k = \arg\max_{\pi} J(\pi, \pi)$. We launch $K$ independent MAPPO runs and keep their final checkpoints,
\begin{equation}
    \Pi_{\text{SP}} = \{\pi_1, \dots, \pi_K\},
    \label{eq:pop_sp}
\end{equation}
so the only source of diversity is the random initialization and the sampling noise of each run. Runs that converge to the same convention leave the population nearly degenerate, and the cooperator then inherits the brittleness of a single self-play agent.

\paragraph{Fictitious Co-Play (FCP).} FCP reuses the same self-play runs and adds their intermediate checkpoints, on the grounds that a cooperator should also handle partners of lower skill. Let $\pi_{k,t}$ denote run $k$ after $t$ gradient steps, with $T$ the final step. FCP retains $C$ checkpoints per run at steps $t_1 < \dots < t_C = T$, which yields
\begin{equation}
    \Pi_{\text{FCP}} = \big\{ \pi_{k,t_c} \big\}_{k = 1, \dots, K}^{c = 1, \dots, C}, \qquad |\Pi_{\text{FCP}}| = KC.
    \label{eq:pop_fcp}
\end{equation}
We use $K = 4$ runs and $C = 3$ checkpoints per run, sampled evenly from the early, middle, and final stages of training.

\paragraph{Maximum Entropy Population-based Training (MEP).} MEP trains the $K$ partners jointly and rewards them for acting differently from one another. Let $\bar{\pi}(a \mid o) = \frac{1}{K} \sum_{k=1}^{K} \pi_k(a \mid o)$ be the mean population policy. Each partner maximizes the team return augmented with the entropy of this mean policy,
\begin{equation}
    \max_{\pi_1, \dots, \pi_K} \; \sum_{k=1}^{K} \mathbb{E}_{\tau \sim (\pi_k, \pi_k)} \left[ \sum_{t} \gamma^t \big( r_t + \eta \, \mathcal{H}\big(\bar{\pi}(\cdot \mid o_t)\big) \big) \right],
    \label{eq:mep}
\end{equation}
where $\eta$ weights the population entropy bonus. We use $\eta = 0.001$ for \environment{LBF} and \environment{PMR}, and $\eta = 0.005$ for \environment{PP} and \environment{overcooked}. This bonus decomposes into an individual and a collective term,
\begin{equation}
    \mathcal{H}(\bar{\pi}) = \underbrace{\frac{1}{K} \sum_{k=1}^{K} \mathcal{H}(\pi_k)}_{\text{stochasticity of each partner}} \; + \; \underbrace{\mathrm{JSD}(\pi_1, \dots, \pi_K)}_{\text{disagreement between partners}},
    \label{eq:mep_decomposition}
\end{equation}
so maximizing it rewards both randomness within a partner and divergence between partners. 

The second stage prioritizes partners that the cooperator performs poorly with by replacing $\operatorname{Uniform}(\Pi)$ in Equation~\ref{eq:pp} with the rank-based prioritized sampling scheme. Specifically, each partner is sampled with probability
\begin{equation}
p(\pi_k) = \frac{\operatorname{rank}\!\left(1 / J(\pi_\phi, \pi_k)\right)^{\beta}}{\sum_{j=1}^{K} \operatorname{rank}\!\left(1 / J(\pi_\phi, \pi_j)\right)^{\beta}},
\label{eq:mep_sampling}
\end{equation}
where $\operatorname{rank}$ assigns 1 to the partner with the highest return and $K$ to the partner with the lowest, and $\beta$ controls the strength of the prioritization. During training, we estimate $J(\pi_\phi, \pi_k)$ as the mean episodic return from the most recent rollouts with $\pi_k$, and sample a partner from $p$ at the start of each episode. \popplay{} uses the same sampling scheme, but with a population of \popgen{}-generated policy code.

\paragraph{CoMeDi.} CoMeDi builds the population one partner at a time and encourages each new partner to adopt a convention that is distinct from those already represented. Suppose $\pi_1, \dots, \pi_{k-1}$ have been trained. For a candidate $\pi$, we identify the existing partner with which it achieves the highest return,
\begin{equation}
j^\star(\pi) = \arg\max_{j < k} \; J(\pi, \pi_j),
\label{eq:comedi_nn}
\end{equation}
and treat this partner as the candidate's nearest existing convention. The new partner then maximizes its own self-play return while minimizing its return when paired with this nearest neighbor,
\begin{equation}
\pi_k = \arg\max_{\pi} \; J(\pi, \pi) \; - \; \lambda_{\text{XP}} \, J\big(\pi, \pi_{j^\star(\pi)}\big) \; + \; \lambda_{\text{MP}} \, J_{\text{MP}}\big(\pi, \pi_{j^\star(\pi)}\big).
\label{eq:comedi}
\end{equation}
The mixed-play return $J_{\text{MP}}$ prevents the candidate from achieving a low paired return simply by failing deliberately. Its rollout initially pairs $\pi$ with $\pi_{j^\star}$, then switches at a random step $t \sim \mathrm{Uniform}(\{0, \dots, T\})$ to using $\pi$ for both agents, and scores only the reward collected after the switch. Maximizing this return encourages $\pi$ to remain competent in states reached when paired with the existing partner. Thus, a low $J(\pi, \pi_{j^\star})$ reflects a genuinely different convention rather than deliberate failure when paired with an unfamiliar partner. We use $\lambda_{\text{XP}} = 0.25$ and $\lambda_{\text{MP}} = 1.0$ from the original implementation.

\paragraph{The CoMeDi cooperator.} CoMeDi also replaces the best response of Equation~\ref{eq:pp} with a multitask objective in the style of QDagger~\citep{agarwal2022reincarnating}. Its convention-aware cooperator $\hat{\pi}$ minimizes
\begin{equation}
    \mathcal{L}(\hat{\pi}, \Pi) = - J(\hat{\pi}, \hat{\pi}) \; - \; \frac{\lambda}{|\Pi|} \sum_{\pi \in \Pi} \; \mathbb{E}_{(o, a) \sim \pi} \big[ \log \hat{\pi}(a \mid o) \big],
    \label{eq:comedi_coop}
\end{equation}
where $(o, a) \sim \pi$ denotes observation-action pairs drawn from self-play rollouts of $\pi$, and $\lambda$ weights the behavior-cloning term. The demonstrations come from a fixed population, so this cooperator only covers the conventions that $\Pi$ already contains.


\subsection{Held-out Evaluation Partners}
\label{app:eval_partners}

The held-out partners are agents that are never seen during training and are paired with cooperators only at evaluation time. We design four partners per environment. Table~\ref{tab:eval_partners} lists the partners for each environment.
\begin{table}[h]
\centering
\footnotesize
\setlength{\tabcolsep}{4pt}
\begin{tabularx}{\linewidth}{l l >{\raggedright\arraybackslash}X}
\toprule
\textbf{Env.} & \textbf{Name} & \textbf{Description} \\
\midrule
LBF
& Closest-Food Greedy & Moves to the nearest food and loads it when possible. \\
& Team-Center Food & Targets food near the center of the two agents. \\
& Teammate-Biased Food & Targets the food cheapest for the teammate to reach. \\
& Synchronized Loading & Waits at its food until the teammate is adjacent. \\
\midrule
PP
& Closest-Prey Greedy & Chases the nearest prey, whether it is a rabbit or a stag. \\
& Teammate-Biased Prey & Targets the prey the teammate is also pressuring. \\
& Rabbit Only & Always pursues the nearest rabbit, never a stag. \\
& Stag Only & Pursues stags only, so it scores only if the partner joins. \\
\midrule
Overcooked
& Fill only & Only fetches onions and loads the pot. \\
& Fetch \& Serve & Only fetches a dish and delivers soup. \\
& Fill \& Start & Loads onions and starts the pot, but never serves. \\
& Wait \& Serve & Waits at the pot until the soup is cooked, then serves. \\
\midrule
PMR
& Closest-Landmark Greedy & Moves toward the currently nearest landmark. \\
& Farthest-Landmark Commitment & Always moves toward the currently farthest landmark. \\
& Teammate-Biased Landmark & Selects the landmark closest to the teammate. \\
& Regroup then Diverge & Approaches the teammate, then heads to a far landmark. \\
\bottomrule
\end{tabularx}
\caption{Held-out evaluation partners for each environment: Level-Based Foraging (LBF),
Predator-Prey (PP), Overcooked, and Point-Mass Rendezvous (PMR).}
\label{tab:eval_partners}
\end{table}
\FloatBarrier

\subsection{Hyperparameters}
\label{app:cooperator}

We train every cooperator with PPO~\citep{schulman2017ppo} on a recurrent actor and critic. We keep the partner policies frozen so that only the cooperator is updated. Table~\ref{tab:cooperator_hparams} reports the hyperparameters. Training populations contain $4$ partners for \popplay{}, Self-Play, MEP, and CoMeDi, and $12$ for FCP, which pools three checkpoints from each of four self-play runs. In the population generation stage, baseline populations are trained with MAPPO~\citep{yu2022surprising}, since their agents learn jointly.
\begin{table}[h]
    \centering
    \caption{Cooperator training hyperparameters. The upper block lists the settings shared by
    every environment and every method. The lower block lists the per-environment settings.}
    \label{tab:cooperator_hparams}
    \small
    \renewcommand{\arraystretch}{1.2}
    \begin{tabular}{ll}
    \toprule
    \multicolumn{2}{c}{\textbf{Shared across all environments}} \\
    \midrule
    Discount factor $\gamma$            & 0.99 \\
    Learning rate                       & $10^{-4}$ \\
    PPO clip $\epsilon$                 & 0.2 \\
    GAE $\lambda$                       & 0.9 \\
    Policy network                      & GRU \\
    Prioritized sampling $\beta$        & 1.0 \\
    \bottomrule
    \end{tabular}

    \vspace{0.6em}

    \begin{tabular}{lccccc}
    \toprule
    \textbf{Parameter} & \textbf{LBF} & \textbf{Overcooked} & \textbf{PMR} & \textbf{PP} \\
    \midrule
    Environment frames      & 4M   & 4M   & 30M   & 80M  \\
    \bottomrule
    \end{tabular}
\end{table}
\FloatBarrier

\subsection{Results Table}
\label{app:cooperator_results}
\begin{table}[!ht]
\centering
\small
\caption{\textbf{Average team reward of each cooperator against evaluation and training partners.} We report reward against held-out \emph{eval} behaviors and against the \emph{train} partners the cooperator saw during training. We report the reward averaged across partners (mean $\pm$ standard deviation over five runs). Best value per environment is in \textbf{bold}.}
\label{tab:cooperator}
\begin{tabular}{ll@{\hskip 12pt}cccc}
\toprule
\textbf{Method} & \textbf{Partners} & PMR & LBF & PP & Overcooked \\
\midrule
\multirow{2}{*}{Self-Play (SP)}
 & Eval  & $1.01 \pm 0.25$ & $0.81 \pm 0.04$ & $1.21 \pm 0.12$ & $63.38 \pm 65.37$ \\
 & Train & $1.60 \pm 0.22$ & $0.78 \pm 0.02$ & $1.36 \pm 0.14$ & $133.79 \pm 151.41$ \\
\midrule
\multirow{2}{*}{FCP}
 & Eval  & $0.93 \pm 0.31$ & $0.80 \pm 0.03$ & $1.27 \pm 0.09$ & $40.53 \pm 52.23$ \\
 & Train & $\mathbf{1.68 \pm 0.30}$ & $0.80 \pm 0.02$ & $1.31 \pm 0.19$ & $157.29 \pm 134.16$ \\
\midrule
\multirow{2}{*}{MEP}
 & Eval  & $1.45 \pm 0.46$ & $0.72 \pm 0.14$ & $1.30 \pm 0.17$ & $33.79 \pm 37.73$ \\
 & Train & $1.32 \pm 0.17$ & $0.65 \pm 0.30$ & $1.31 \pm 0.32$ & $126.05 \pm 104.18$ \\
\midrule
\multirow{2}{*}{CoMeDi}
 & Eval  & $1.56 \pm 0.39$ & $0.59 \pm 0.09$ & $1.02 \pm 0.13$ & $86.33 \pm 31.48$ \\
 & Train & $1.51 \pm 0.15$ & $0.48 \pm 0.14$ & $0.88 \pm 0.13$ & $\mathbf{312.96 \pm 75.00}$ \\
\midrule
\multirow{2}{*}{\popplay{} (ours)}
 & Eval  & $\mathbf{1.70 \pm 0.56}$ & $\mathbf{0.87 \pm 0.07}$ & $\mathbf{1.59 \pm 0.08}$ & $\mathbf{120.88 \pm 59.25}$ \\
 & Train & $1.64 \pm 0.45$ & $\mathbf{0.94 \pm 0.03}$ & $\mathbf{1.71 \pm 0.29}$ & $211.66 \pm 122.13$ \\
\bottomrule
\end{tabular}
\end{table}
Table~\ref{tab:cooperator} reports the per-environment numbers behind Figure~\ref{fig:zs_cooperation}. We report two different setups: one against the held-out eval partners and the other against the train partners. \popplay{} attains the highest eval reward in every environment. The reward difference between eval and train partners measures how much of its reward is lost due to out-of-distribution partners.

Table~\ref{tab:coop_results_heldout_partners} breaks the eval row of Table~\ref{tab:cooperator} down to the individual partner. We read each row as the partner picking the cooperator it scores highest with. Fourteen of the sixteen partners picked \popplay{} as the most cooperative agent, and the remaining two, both in \environment{overcooked}, picked CoMeDi.
\begin{table}[!ht]
\centering
\small
\setlength{\tabcolsep}{4pt}
\caption{\textbf{Team reward against each held-out partner.} We report mean $\pm$ standard error of the team reward across 5 runs. For every partner, \textbf{bold} marks the best cooperator and every method whose error bar overlaps it. The final row counts the partners for which a method attains the highest average.}
\label{tab:coop_results_heldout_partners}
\begin{tabular}{@{}lrrrrr@{}}
\toprule
Held-out partner & \popplay{} (ours) & SP & FCP & MEP & CoMeDi \\
\midrule
\multicolumn{6}{@{}l}{\emph{Point-Mass Rendezvous (PMR)}} \\
\quad Closest-Landmark Greedy & $\mathbf{1.58}_{\pm 0.26}$ & $0.82_{\pm 0.08}$ & $0.92_{\pm 0.16}$ & $1.01_{\pm 0.02}$ & $\mathbf{1.46}_{\pm 0.21}$ \\
\quad Farthest-Landmark Commitment & $\mathbf{1.60}_{\pm 0.40}$ & $\mathbf{0.80}_{\pm 0.49}$ & $0.40_{\pm 0.40}$ & $\mathbf{1.20}_{\pm 0.49}$ & $\mathbf{1.45}_{\pm 0.22}$ \\
\quad Teammate-Biased Landmark & $\mathbf{2.00}_{\pm 0.00}$ & $\mathbf{2.00}_{\pm 0.00}$ & $\mathbf{2.00}_{\pm 0.00}$ & $\mathbf{1.99}_{\pm 0.01}$ & $1.93_{\pm 0.04}$ \\
\quad Regroup then Diverge & $\mathbf{1.60}_{\pm 0.40}$ & $0.40_{\pm 0.40}$ & $0.40_{\pm 0.40}$ & $\mathbf{1.60}_{\pm 0.40}$ & $\mathbf{1.42}_{\pm 0.23}$ \\
\addlinespace
\multicolumn{6}{@{}l}{\emph{Level-Based Foraging (LBF)}} \\
\quad Closest-Food Greedy & $\mathbf{0.76}_{\pm 0.07}$ & $\mathbf{0.72}_{\pm 0.05}$ & $\mathbf{0.73}_{\pm 0.04}$ & $\mathbf{0.71}_{\pm 0.07}$ & $0.49_{\pm 0.03}$ \\
\quad Team-Center Food & $\mathbf{0.90}_{\pm 0.02}$ & $0.82_{\pm 0.02}$ & $0.81_{\pm 0.01}$ & $0.71_{\pm 0.06}$ & $0.62_{\pm 0.04}$ \\
\quad Teammate-Biased Food & $\mathbf{0.92}_{\pm 0.01}$ & $0.87_{\pm 0.01}$ & $0.86_{\pm 0.02}$ & $0.77_{\pm 0.05}$ & $0.66_{\pm 0.04}$ \\
\quad Synchronized Loading & $\mathbf{0.90}_{\pm 0.02}$ & $0.84_{\pm 0.02}$ & $0.82_{\pm 0.01}$ & $0.70_{\pm 0.07}$ & $0.60_{\pm 0.04}$ \\
\addlinespace
\multicolumn{6}{@{}l}{\emph{Predator-Prey (PP)}} \\
\quad Closest-Prey Greedy & $\mathbf{1.35}_{\pm 0.06}$ & $1.06_{\pm 0.06}$ & $\mathbf{1.28}_{\pm 0.05}$ & $1.16_{\pm 0.10}$ & $1.01_{\pm 0.03}$ \\
\quad Teammate-Biased Prey & $\mathbf{1.81}_{\pm 0.12}$ & $1.43_{\pm 0.07}$ & $1.39_{\pm 0.05}$ & $1.48_{\pm 0.16}$ & $1.20_{\pm 0.11}$ \\
\quad Rabbit Only & $\mathbf{1.38}_{\pm 0.06}$ & $1.21_{\pm 0.04}$ & $1.21_{\pm 0.02}$ & $1.05_{\pm 0.02}$ & $1.24_{\pm 0.03}$ \\
\quad Stag Only & $\mathbf{1.81}_{\pm 0.11}$ & $1.15_{\pm 0.08}$ & $1.22_{\pm 0.13}$ & $1.53_{\pm 0.14}$ & $0.64_{\pm 0.09}$ \\
\addlinespace
\multicolumn{6}{@{}l}{\emph{Overcooked}} \\
\quad Fill only & $\mathbf{36}_{\pm 18}$ & $12_{\pm 0}$ & $12_{\pm 0}$ & $\mathbf{17}_{\pm 4}$ & $11_{\pm 0}$ \\
\quad Fetch \& Serve & $53_{\pm 29}$ & $0_{\pm 0}$ & $0_{\pm 0}$ & $2_{\pm 2}$ & $\mathbf{141}_{\pm 18}$ \\
\quad Fill \& Start & $\mathbf{341}_{\pm 66}$ & $\mathbf{242}_{\pm 117}$ & $150_{\pm 93}$ & $113_{\pm 70}$ & $81_{\pm 44}$ \\
\quad Wait \& Serve & $53_{\pm 29}$ & $0_{\pm 0}$ & $0_{\pm 0}$ & $2_{\pm 2}$ & $\mathbf{112}_{\pm 20}$ \\
\midrule
Best cooperator & $\mathbf{14}$/16 & $0$/16 & $0$/16 & $0$/16 & $2$/16 \\
\bottomrule
\end{tabular}
\end{table}


\FloatBarrier

\subsection{Mode Collapse in Self-Play Populations}
\label{app:sp_collapse}

Self-Play populations built from independent seeds often collapse to a few conventions. To show this, we pair each self-play policy with itself and record which strategy the pair follows. In \environment{PMR}, different seeds can converge to different landmarks, but each policy commits to a single landmark, and a population of four covers only $2.8$ of the four landmarks on average (Table~\ref{tab:sp_pmr_landmarks}). Since Self-Play has no mechanism that promotes diversity across seeds, several seeds can settle on the same landmark, and all four policies in Run~5 settle on the same one. In \environment{overcooked}, self-play pairs mostly put onions into the pot and rarely fetch a dish or scoop soup (Table~\ref{tab:sp_overcooked_stages}). Cooperators trained against these populations therefore rarely see partners that take the other roles. FCP suffers from the same collapse, since its checkpoints follow the same semantic behaviors as the self-play runs and differ only in skill level.
\begin{table}[!ht]
\centering
\small
\caption{\textbf{Self-Play in Point-Mass Rendezvous.} Each column is one self-play policy paired with itself. We run 5 self-play experiments, each producing 4 policies from different seeds, and evaluate each policy for 20 episodes. Rows give the percentage of episodes in which the pair meets at each landmark, and the largest in each column is in bold. Counting each policy's most frequent landmark, only Run~1 covers all four landmarks, and all four policies in Run~5 settle on West, since Self-Play has no principled way of promoting diversity across seeds.}
\label{tab:sp_pmr_landmarks}
\scriptsize
\setlength{\tabcolsep}{2.5pt}
\setlength{\aboverulesep}{0pt}
\setlength{\belowrulesep}{0pt}
\renewcommand{\arraystretch}{1.25}
\begin{tabularx}{\textwidth}{l|*{4}{>{\centering\arraybackslash}X}|*{4}{>{\centering\arraybackslash}X}|*{4}{>{\centering\arraybackslash}X}|*{4}{>{\centering\arraybackslash}X}|*{4}{>{\centering\arraybackslash}X}}
\toprule
Landmark (\%) & \multicolumn{4}{c}{Run~1} & \multicolumn{4}{c}{Run~2} & \multicolumn{4}{c}{Run~3} & \multicolumn{4}{c}{Run~4} & \multicolumn{4}{c}{Run~5} \\
\cmidrule(lr){2-5} \cmidrule(lr){6-9} \cmidrule(lr){10-13} \cmidrule(lr){14-17} \cmidrule(lr){18-21}
 & 1 & 2 & 3 & 4 & 1 & 2 & 3 & 4 & 1 & 2 & 3 & 4 & 1 & 2 & 3 & 4 & 1 & 2 & 3 & 4 \\
\midrule
East & 0 & 0 & 0 & \textbf{100} & 35 & \textbf{100} & 0 & 0 & 0 & 0 & 0 & 0 & \textbf{85} & 0 & 0 & \textbf{100} & 0 & 0 & 0 & 0 \\
North & \textbf{100} & 15 & 0 & 0 & 0 & 0 & \textbf{75} & 0 & 0 & 0 & \textbf{100} & 0 & 15 & 0 & 30 & 0 & 0 & 0 & 0 & 0 \\
West & 0 & 0 & \textbf{75} & 0 & 0 & 0 & 0 & 0 & \textbf{100} & 0 & 0 & 0 & 0 & 0 & \textbf{70} & 0 & \textbf{100} & \textbf{100} & \textbf{70} & \textbf{100} \\
South & 0 & \textbf{85} & 25 & 0 & \textbf{65} & 0 & 25 & \textbf{100} & 0 & \textbf{100} & 0 & \textbf{100} & 0 & \textbf{100} & 0 & 0 & 0 & 0 & 30 & 0 \\
\bottomrule
\end{tabularx}
\end{table}

\begin{table}[!ht]
\centering
\small
\caption{\textbf{Self-Play in Overcooked.} Each column is one self-play policy paired with itself. We run 5 self-play experiments, each producing 4 policies from different seeds, and evaluate each policy for 20 episodes. Rows give each pipeline stage's share of all stage actions performed by both agents (\%), and the largest in each column is in bold.}
\label{tab:sp_overcooked_stages}
\scriptsize
\setlength{\tabcolsep}{2pt}
\setlength{\aboverulesep}{0pt}
\setlength{\belowrulesep}{0pt}
\renewcommand{\arraystretch}{1.25}
\begin{tabularx}{\textwidth}{l|*{4}{>{\centering\arraybackslash}X}|*{4}{>{\centering\arraybackslash}X}|*{4}{>{\centering\arraybackslash}X}|*{4}{>{\centering\arraybackslash}X}|*{4}{>{\centering\arraybackslash}X}|r}
\toprule
Stage (\%) & \multicolumn{4}{c}{Run~1} & \multicolumn{4}{c}{Run~2} & \multicolumn{4}{c}{Run~3} & \multicolumn{4}{c}{Run~4} & \multicolumn{4}{c}{Run~5} &  \\
\cmidrule(lr){2-5} \cmidrule(lr){6-9} \cmidrule(lr){10-13} \cmidrule(lr){14-17} \cmidrule(lr){18-21}
 & 1 & 2 & 3 & 4 & 1 & 2 & 3 & 4 & 1 & 2 & 3 & 4 & 1 & 2 & 3 & 4 & 1 & 2 & 3 & 4 & Mean \\
\midrule
Fetch onion & \textbf{37} & \textbf{40} & \textbf{51} & \textbf{38} & \textbf{58} & \textbf{51} & \textbf{40} & \textbf{54} & \textbf{41} & \textbf{49} & \textbf{35} & \textbf{46} & \textbf{43} & \textbf{42} & \textbf{57} & \textbf{34} & \textbf{51} & \textbf{47} & \textbf{35} & \textbf{39} & 44 \\
Onion$\to$pot & 31 & 36 & 38 & 35 & 38 & 36 & 35 & 39 & 33 & 28 & 33 & 45 & 33 & 35 & 34 & 34 & 38 & 42 & 35 & 36 & 36 \\
Start pot & 9 & 12 & 0 & 12 & 0 & 0 & 12 & 0 & 8 & 8 & 11 & 0 & 11 & 10 & 0 & 10 & 0 & 0 & 12 & 12 & 6 \\
Fetch dish & 11 & 12 & 11 & 8 & 4 & 13 & 13 & 7 & 11 & 8 & 11 & 10 & 11 & 13 & 9 & 10 & 10 & 11 & 10 & 12 & 10 \\
Scoop soup & 11 & 0 & 0 & 8 & 0 & 0 & 0 & 0 & 8 & 8 & 11 & 0 & 2 & 0 & 0 & 10 & 0 & 0 & 10 & 1 & 3 \\
Serve & 0 & 0 & 0 & 0 & 0 & 0 & 0 & 0 & 0 & 0 & 0 & 0 & 0 & 0 & 0 & 0 & 0 & 0 & 0 & 0 & 0 \\
\bottomrule
\end{tabularx}
\end{table}

\FloatBarrier



\subsection{Task Event Counts in \environment{overcooked}}
\label{app:oc_events}

Table~\ref{tab:oc_events_full} breaks down the subtask completion shares in \environment{overcooked} from Figure~\ref{tab:oc_events} by held-out partner, reporting the partner's counts alongside those of the cooperator. \popplay{} shifts toward subtasks left open by the partner, whereas Self-Play, FCP, and MEP remain focused on plating and show little activity when the partner already occupies those subtasks.
\begin{table}[h]
\centering
\small
\renewcommand{\arraystretch}{1.15}
\caption{\textbf{Cooperator task events per held-out partner in \environment{overcooked}.} Per-partner
breakdown of Figure~\ref{tab:oc_events}. Fill \& Start and Fill only load the pot but never serve. Fetch \& Serve and Wait \& Serve plate and serve but never fetch an
onion. Counts are averaged per episode over 50 episodes.}
\label{tab:oc_events_full}
\begin{tabular}{ll@{\hskip 12pt}ccccc}
\toprule
& & \multicolumn{5}{c}{\textbf{Cooperator}} \\
\cmidrule(lr){3-7}
\textbf{Method} & \textbf{Partner} & Onion $\rightarrow$ pot & Start pot & Fetch dish & Scoop soup & Serve soup \\
\midrule
\multirow{4}{*}{\popplay{}}
 & Fill \& Start   & $0.23$ & $0.22$ & $10.62$ & $10.35$ & $8.14$ \\
 & Fill only       & $0.23$ & $0.78$ & $2.34$ & $0.74$ & $0.72$ \\
 & Fetch \& Serve  & $5.18$ & $4.85$ & $0.36$ & $0.02$ & $0.00$ \\
 & Wait \& Serve   & $5.20$ & $4.85$ & $0.41$ & $0.03$ & $0.00$ \\
\midrule
\multirow{4}{*}{Self-Play}
 & Fill \& Start   & $0.00$ & $0.00$ & $7.96$ & $7.16$ & $5.55$ \\
 & Fill only       & $0.00$ & $0.00$ & $22.47$ & $0.00$ & $0.00$ \\
 & Fetch \& Serve  & $0.00$ & $0.00$ & $0.05$ & $0.00$ & $0.00$ \\
 & Wait \& Serve   & $0.00$ & $0.00$ & $0.05$ & $0.00$ & $0.00$ \\
\midrule
\multirow{4}{*}{FCP}
 & Fill \& Start   & $0.00$ & $0.00$ & $5.51$ & $4.84$ & $2.96$ \\
 & Fill only       & $0.00$ & $0.00$ & $1.28$ & $0.00$ & $0.00$ \\
 & Fetch \& Serve  & $0.00$ & $0.00$ & $0.27$ & $0.00$ & $0.00$ \\
 & Wait \& Serve   & $0.00$ & $0.00$ & $0.27$ & $0.00$ & $0.00$ \\
\midrule
\multirow{4}{*}{MEP}
 & Fill \& Start   & $0.03$ & $0.32$ & $4.63$ & $3.76$ & $2.05$ \\
 & Fill only       & $0.03$ & $0.37$ & $1.24$ & $0.37$ & $0.08$ \\
 & Fetch \& Serve  & $0.32$ & $0.18$ & $0.23$ & $0.04$ & $0.00$ \\
 & Wait \& Serve   & $0.31$ & $0.18$ & $0.23$ & $0.05$ & $0.00$ \\
\midrule
\multirow{4}{*}{CoMeDi}
 & Fill \& Start   & $0.59$ & $0.06$ & $2.90$ & $1.93$ & $1.91$ \\
 & Fill only       & $0.54$ & $0.07$ & $1.89$ & $0.03$ & $0.03$ \\
 & Fetch \& Serve  & $12.83$ & $4.16$ & $0.63$ & $0.00$ & $0.00$ \\
 & Wait \& Serve   & $10.31$ & $3.30$ & $0.71$ & $0.02$ & $0.01$ \\
\bottomrule
\end{tabular}%
\end{table}

\FloatBarrier

\section{Diverse Behavior Generation}
\label{app:behavior_generation}

In this section, we extensively ablate methods to generate diverse text while retaining maximum quality~\citep{pugh2016qualitydiversity}. We include two additional environments for completeness, as the following experiments were cheaper to run on them\footnote{For both, the main issue is rollout completion time, which makes them hard to include in the rest of the paper because training becomes quite long. Football rollouts can be quite long before anything concrete happens, and SMACv2 runs a slow StarCraft backend, which was hard to optimize for fast training.}. 

\paragraph{Football~\citep{bettini2024benchmarl}.} The Football environment introduces a team setting in which agents must maintain positions relative to the ball, goals, and teammates while dynamically switching between offensive and defensive roles.

\paragraph{SMACv2~\citep{ellis2023smacv2improvedbenchmarkcooperative}.} SMACv2 consists of small-scale StarCraft multi-agent combat scenarios in which allied units coordinate against a stronger enemy force under partial observability. In our setting, two allied Marines face four enemy Marines, so success depends on tactical movement, focus fire, spacing, retreat timing, and target selection.



\begin{figure}[t]
    \centering
    \includegraphics[width=0.6\linewidth]{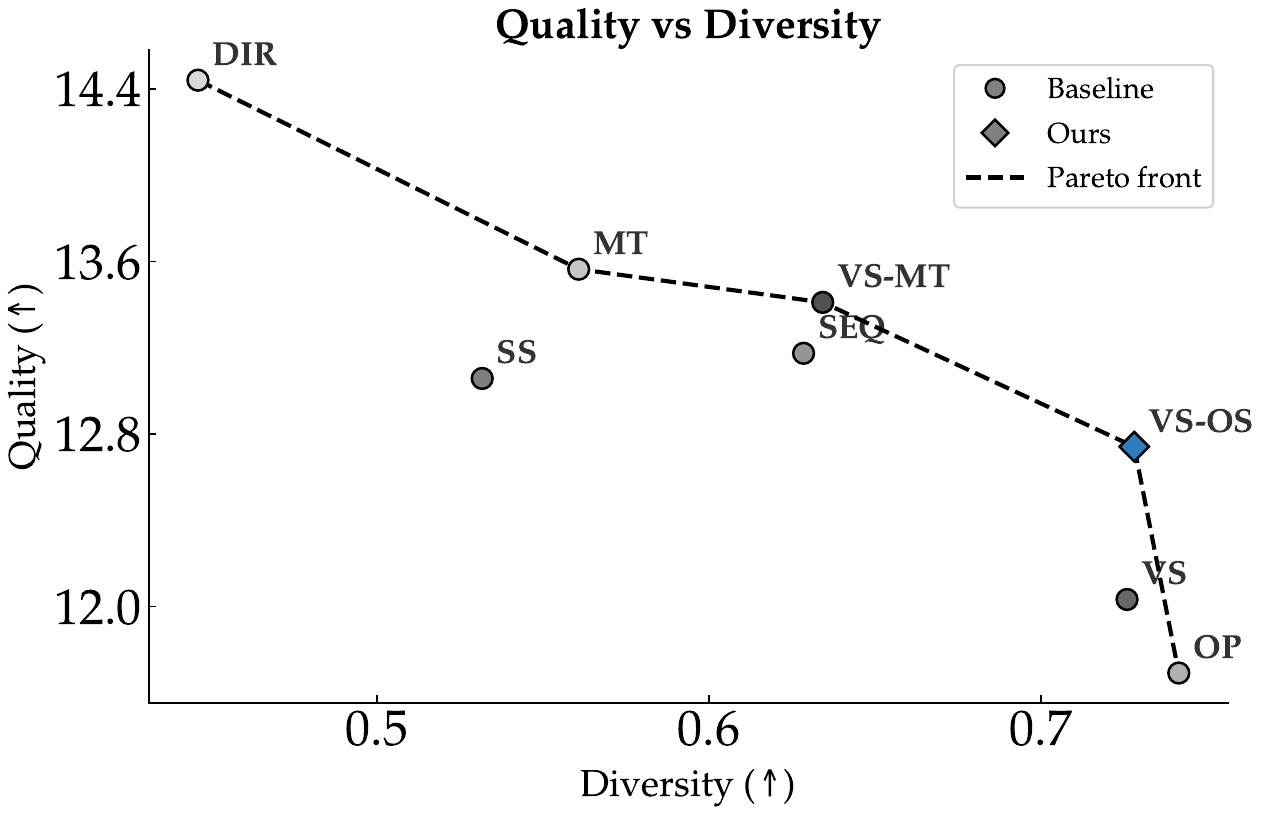}
    \caption{\textbf{Quality and diversity of all behavior generation methods.} Quality and diversity of a generated set of 30 behaviors, averaged across the four environments, for every method in Table~\ref{tab:prompting_methods}. VS-OS is the method used in \popgen{}.}
    \label{fig:behavior_generation_all}
\end{figure}

\paragraph{Diverse text generation methods.}
\begin{table}[ht]
\caption{\textbf{Comparison of prompting methods for generating $N$ behaviors.} $k$ is the number of candidates per LLM call (in our experiments $k=5$). $y_i$ denotes the $i$-th generated candidate, $\hat{p}_i$ its verbalized probability, and $\pi(\cdot|x)$ the LLM's output distribution conditioned on prompt $x$. For Multi-Turn and VS-Multi, $h_{i-1}$ denotes the conversation history up to turn $i - 1$, and $t$ the turn index.}
\label{tab:prompting_methods}
\renewcommand{\arraystretch}{1.6}
\resizebox{\textwidth}{!}{%
\begin{tabular}{@{}llccll@{}}
\toprule
\textbf{Method} & \textbf{LLM Calls} & \textbf{Candidates} & \textbf{Turns} & \textbf{Prompt Example} & \textbf{Definition} \\
\midrule
\multicolumn{6}{l}{\textit{\textbf{1. Instance-level Prompt}}} \\
\hdashline
\quad Direct (DIR) & $N$ & 1 & 1 & ``Think step-by-step, generate a football player behavior'' & $y_i \sim \pi(y|x_\text{CoT})$ \\
\midrule
\multicolumn{6}{l}{\textit{\textbf{2. List-level Prompt}}} \\
\hdashline
\quad One-Pass (OP) & $1$ & $N$ & 1 & ``Generate 30 football player behaviors'' & $(y_1, \ldots, y_N) \sim \pi(y_1, \ldots, y_N | x_\text{OP})$ \\
\hdashline
\quad Sequence (SEQ) & $\lceil N/k \rceil$ & $k$ & 1 & ``Generate 5 football player behaviors'' & $(y_1, \ldots, y_k) \sim \pi(y_1, \ldots, y_k | x_\text{seq})$ \\
\hdashline
\quad Multi-Turn (MT) & $N$ & 1 & $N$ &
\makecell[l]{Turn 1: ``Generate a football player behavior'' \\ Turn 2+: ``Generate another football player behavior''} &
$y_i \sim \pi(y | x_\text{multi}, h_{i-1})$ \\
\hdashline
\quad SimpleStrat (SS)~\citep{wong2024simplestrat} & $\lceil N/k \rceil$ & $k$ & 1 &
\makecell[l]{``Generate a football player behavior following \\ these characteristics: [strata]''} &
$y_i \sim \pi(y \mid x_{\text{strat}_i})$,\; $\text{strat}_i \sim \hat{P}_\text{strata}$ \\
\midrule
\multicolumn{6}{l}{\textit{\textbf{3. Distribution-level Prompt}}} \\
\hdashline
\quad Verbalized-Sampling (VS)~\citep{zhang2025verbalizedsampling} & $\lceil N/k \rceil$ & $k$ & 1 &
\makecell[l]{``Think step-by-step, generate 5 football \\ player behaviors with probabilities''} &
$(y_1, \hat{p}_1), \ldots, (y_k, \hat{p}_k) \sim \pi(\cdot | x_\text{VS-CoT})$ \\
\hdashline
\quad VS-MT~\citep{zhang2025verbalizedsampling} & $\lceil N/k \rceil$ & $k$ & $\lceil N/k \rceil$ &
\makecell[l]{Turn 1: ``Generate 5 player behaviors with probabilities'' \\ Turn 2+: ``Generate 5 more with probabilities''} &
$\begin{pmatrix} (y_1^{(1)}, \hat{p}_1^{(1)}), \ldots, (y_k^{(t)}, \hat{p}_k^{(t)}) \\ \sim \pi(\cdot | x_\text{VS}, h_{t-1}) \end{pmatrix}$ \\
\hdashline
\quad VS-OS (ours) & $\lceil m/k \rceil + 1$ & $k$ & 1 &
\makecell[l]{(VS) ``Generate $m{>}N$ player behaviors with probs'' \\ (Merge) ``Combine into $N$ best behaviors''} &
\makecell[l]{$\{(y_i, \hat{p}_i)\}_{i=1}^{m} \sim \pi(\cdot \mid x_\text{VS}),\; m > N$ \\ $\tilde{y}_{1:N} = f_\pi\!\left(\{(y_i,\hat{p}_i)\}_{i=1}^{m}\right)$} \\
\bottomrule
\end{tabular}%
}
\end{table}
Table~\ref{tab:prompting_methods} presents the generation methods we evaluate, organized by the level at which diversity is encouraged. \emph{Instance-level} methods (Direct) generate one behavior per independent LLM call with chain-of-thought reasoning~\citep{wei2022chainofthought}, relying solely on sampling stochasticity for diversity. \emph{List-level} methods ask the model to produce multiple behaviors in a single output: One-Pass generates all $N$ behaviors at once, Sequence generates them in batches of $k$, and Multi-Turn uses conversation history so that each new behavior is conditioned on previously generated ones. SimpleStrat~\citep{wong2024simplestrat} first asks the LLM to identify meaningful diversity axes (strata) for the environment, then samples behaviors conditioned on randomly selected strata combinations. \emph{Distribution-level} methods build on Verbalized Sampling (VS)~\citep{zhang2025verbalizedsampling}, which prompts the model to output behaviors together with verbalized probabilities, encouraging it to sample from the tails of its distribution rather than collapsing to the mode. VS-MT extends this across conversation turns, accumulating history. Finally, we propose \emph{VS-Oversample} (VS-OS), which generates $m > N$ candidate behaviors with verbalized probabilities, then uses a second LLM call to merge and deduplicate them into the $N$ best behaviors. This two-step approach decouples diversity exploration (oversampling) from quality control (merging), allowing broader coverage without sacrificing coherence. Figure~\ref{fig:behavior_generation_all} compares all methods, of which Section~\ref{sec:diverse_behaviors} reports a subset: \popgen{} corresponds to VS-OS, SemDiv to MT, and DIR and VS to themselves.

\subsection{Oversampling to Preserve Quality in Behavior Generation}

\begin{table}[h]
  \centering
  \caption{\textbf{Quality and diversity of oversampling variants across environments.}
           Diversity is measured as average pairwise embedding distance; quality as total
           score. Best value per column in \textbf{bold}. The Sequential $\times$2 variant
           --- referred to as \texttt{VS-OS} throughout the paper --- achieves the best
           diversity--quality trade-off overall.}
  \label{tab:quality_diversity_os}
  \renewcommand{\arraystretch}{1.15}
  \resizebox{\linewidth}{!}{%
  \begin{tabular}{ll cccccc >{\columncolor{gray!15}}c cccccc >{\columncolor{gray!15}}c}
  \toprule
  & & \multicolumn{7}{c}{\textbf{Diversity} ($\uparrow$)}
    & \multicolumn{7}{c}{\textbf{Quality} ($\uparrow$)} \\
  \cmidrule(lr){3-9} \cmidrule(lr){10-16}
  \textbf{Type} & \textbf{Ratio}
    & \environment{football} & \environment{lbf} & \environment{pmr} & \environment{pp} & \environment{smacv2} & \environment{overcooked} & avg
    & \environment{football} & \environment{lbf} & \environment{pmr} & \environment{pp} & \environment{smacv2} & \environment{overcooked} & avg \\
  \midrule
  Baseline (VS) & $\times$1
    & 0.730          & \textbf{0.726} & 0.636          & 0.691          & \textbf{0.756} & 0.612          & 0.692
    & 12.77          & 11.16          & 13.52          & 12.20          & 12.00          & \textbf{13.32} & 12.50 \\
  \midrule
  \multirow{2}{*}{One Pass}
    & $\times$2 & 0.737          & 0.710 & 0.645          & 0.725 & 0.752 & 0.634          & \textbf{0.701}
               & 12.47          & 12.24 & 13.78          & 11.90 & 12.44 & 13.30          & 12.69 \\
    & $\times$4 & 0.729          & 0.684 & 0.644          & 0.718 & 0.747 & \textbf{0.684} & \textbf{0.701}
               & 12.50          & 12.18 & 13.84          & 13.04 & 12.57 & 13.20          & 12.89 \\
  \midrule
  \multirow{2}{*}{Independent}
    & $\times$2 & 0.737 & 0.654          & 0.637          & 0.735 & 0.738 & 0.616 & 0.686
               & 12.44 & 11.97          & 13.84          & 12.93 & 12.20 & 13.26 & 12.77 \\
    & $\times$4 & 0.726 & 0.672          & 0.643          & 0.721 & 0.749 & 0.614 & 0.688
               & \textbf{13.01} & \textbf{12.57} & \textbf{13.88} & 12.77 & 12.11 & 13.10 & 12.91 \\
  \midrule
  \multirow{2}{*}{Sequential}
    & $\times$2 & 0.733          & 0.703 & \textbf{0.649} & \textbf{0.737} & 0.740 & 0.613 & 0.696
               & \textbf{13.01} & 11.97 & 13.78          & \textbf{13.14} & \textbf{12.87} & 13.20 & \textbf{13.00} \\
    & $\times$4 & \textbf{0.738} & 0.685 & 0.640          & 0.719          & 0.694 & 0.594 & 0.678
               & 12.84          & 12.36 & 13.74          & 12.66          & 12.80 & 13.22 & 12.94 \\
  \bottomrule
  \end{tabular}%
  }
\end{table}

\paragraph{Oversampling ablation.}
Table~\ref{tab:quality_diversity_os} ablates how we oversample, comparing One-Pass, Independent, and Sequential strategies at ratios $\times$2 and $\times$4 against vanilla VS ($\times$1). All variants improve quality over the VS baseline while maintaining comparable diversity, confirming that the merging step filters low-quality candidates without collapsing diversity. Sequential $\times$2 achieves the best overall balance, reaching the best average quality (13.00) while staying within $0.005$ of the highest average diversity (0.696 against 0.701), and we adopt it as the default VS-OS configuration throughout the paper. Increasing the ratio to $\times$4 leaves diversity essentially unchanged for One-Pass and Independent while lowering it for Sequential, suggesting that an overly large candidate pool gives the merging step too much freedom to prune unusual behaviors in favor of safe ones.

\begin{figure}[t]
    \centering
    \subfloat[Detailed diversity metrics: minimum pairwise embedding distance, embedding space coverage, and lexical dissimilarity (1 $-$ Rouge-L).\label{fig:bg_diversity}]{%
        \includegraphics[width=\linewidth]{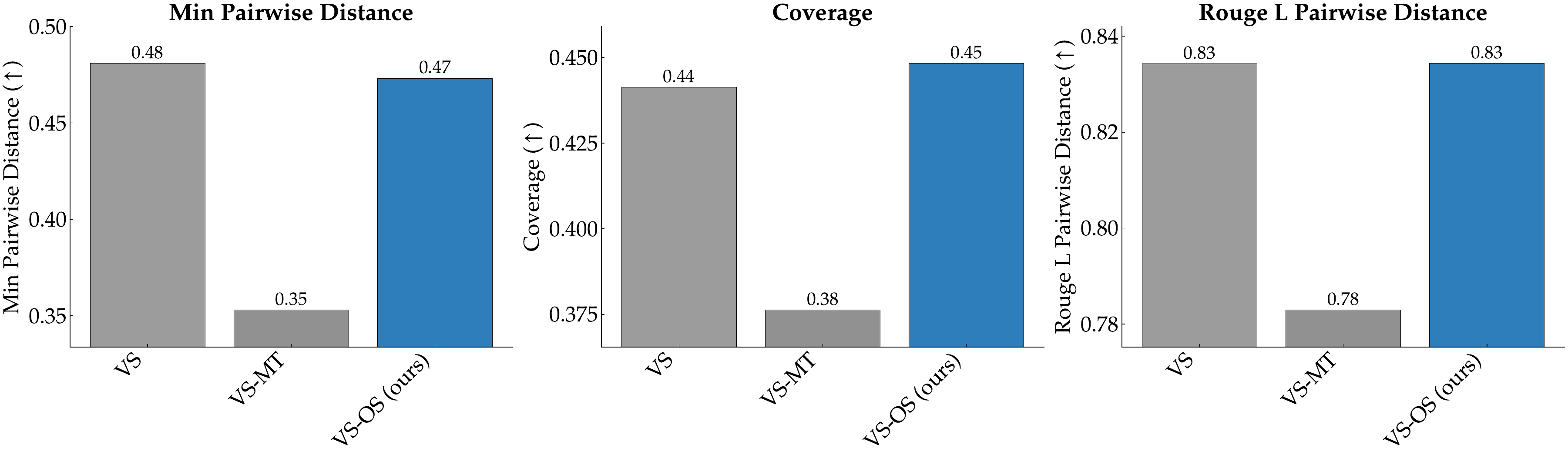}%
    }
    \\
    \subfloat[Scaling VS-OS: generation latency in seconds (left); quality improvement $\Delta$ when scaling the generator from Flash to Pro (center); diversity gain $\Delta$ from the same scaling (right).\label{fig:bg_deltas}]{%
        \includegraphics[width=\linewidth]{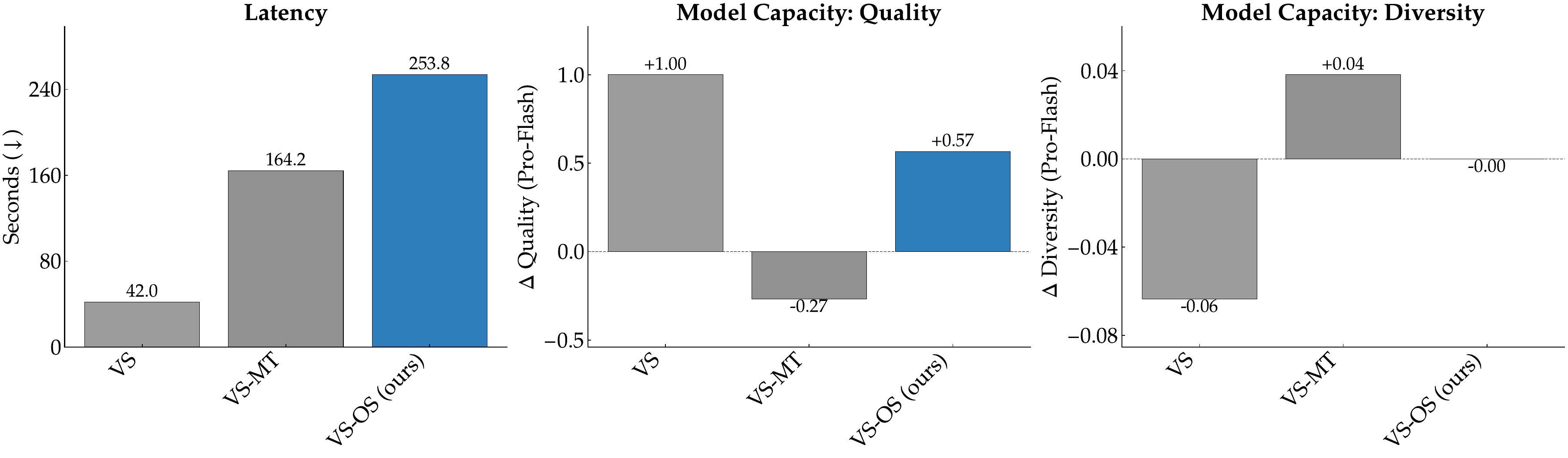}%
    }
    \caption{\textbf{Diversity breakdown and scaling properties of VS variants.} (a)~VS-MT sacrifices worst-case diversity (minimum pairwise distance drops to 0.35) and coverage (0.38) relative to single-pass VS (0.48, 0.44): iterative refinement tends to homogenize outputs. VS-OS recovers both metrics (0.47, 0.45) by decoupling exploration from selection---oversampling casts a wide net, then a dedicated merging step curates without collapsing variety. Lexical dissimilarity is comparable across all three. (b)~The decoupled design of VS-OS trades additional latency for better scaling behavior. When moving from Flash to Pro, VS-MT gains quality ($+1.00$) but \emph{loses} diversity ($-0.06$), confirming that multi-turn refinement with a stronger model amplifies the homogenization effect. VS-OS, by contrast, translates the stronger model into gains on \emph{both} axes ($+0.57$ quality, $+0.04$ diversity): because exploration and selection are separate stages, a more capable model improves the quality of each stage independently without one undermining the other.}
    \label{fig:bg_combined}
\end{figure}
\paragraph{Diversity analysis and scaling of VS-OS.}
Figure~\ref{fig:bg_combined} refines the comparison between VS-OS and its single-pass counterpart. Both achieve comparable lexical dissimilarity, but they diverge on embedding-level measures: VS-OS recovers the worst-case (minimum pairwise) distance and coverage that multi-turn VS sacrifices (Figure~\ref{fig:bg_diversity}), indicating that oversample-and-merge fills gaps a single pass leaves unexplored rather than duplicating its distribution. Figure~\ref{fig:bg_deltas} shows why this matters at scale: when upgrading the generator from Flash to Pro, multi-turn VS gains quality ($+1.00$) but \emph{loses} diversity ($-0.06$), whereas VS-OS translates the same upgrade into gains on \emph{both} axes ($+0.57$ quality, $+0.04$ diversity). Because exploration and selection are separate stages, a more capable model improves each independently.

\paragraph{Transfer of findings from other domains.}
Figure~\ref{fig:behavior_generation_all} includes generation methods that were tested in other domains. SimpleStrat (13.07 quality, 0.532 diversity) falls below the Pareto front, indicating that stratified partitioning, while effective in general-knowledge QA~\citep{wong2024simplestrat}, does not transfer well to generating valid agent strategies where identifying meaningful diversity axes is itself hard. In contrast, the core findings of Verbalized Sampling for creative writing carry over: prompting the model to verbalize a distribution of open-ended responses consistently improves diversity across all VS variants. A consistent secondary pattern is that multi-turn interaction improves quality at the cost of diversity, as conversation history anchors subsequent generations around themes established in earlier turns.

\section{Modeling Behaviors in Agents with LLMs}
\label{app:b2t}

In its second stage, \popgen{} turns each textual behavior description into an executable \textit{policy} that can act in the environment (see Figure~\ref{fig:popgen}). These policies must adhere to their behaviors (i.e., act according to their textual descriptions), otherwise the cooperator would be taught to cooperate with noisy partners, leading to arbitrary conventions that could harm cooperation with real partners. This section ablates this part. We report agent adherence, as introduced in Section~\ref{sec:method_agents}, on two splits: a \emph{dev} set on which the LLM optimizes during refinement, and a \emph{test} set of 10 environment initializations never observed by the refinement loop. The former evaluates how well the LLM refines policies across attempts, while the latter controls for overfitting.

\subsection{Behavior Modeling Methods}
\label{app:b2a_methods}

\begin{table}[t]
  \centering
  \caption{\textbf{Behavior modeling methods.} Each method specifies a distinct mechanism for instantiating a target behavior $b$ in an agent policy $\pi$. $o$ is the observation of the environment, and $a$ is the sampled action.}
  \label{tab:b2a_methods}
  \renewcommand{\arraystretch}{1.4}
  \resizebox{\linewidth}{!}{%
  \begin{tabular}{llp{9cm}}
  \toprule
  \textbf{Name} & \textbf{Abbr.} & \textbf{Description} \\
  \midrule
  LLM-Embodied Agent & LLM-A & Uses an LLM directly as the decision-making agent, conditioned on textualized observations and a behavior specification: $\pi_\text{LLM-A} \sim \text{LLM}(o, b)$ \\
  Reward Function & RF & An LLM generates a reward function $r_b: (o, a) \rightarrow \mathbb{R}$, whose optimization via reinforcement learning yields a neural network policy $\pi_\text{RF}$ exhibiting the target behavior. \\
  Policy Code & PC & An LLM synthesizes an executable policy $f_b: o \rightarrow a$ that directly maps observations to actions, serving as the policy. \\
  \bottomrule
  \end{tabular}%
  }
\end{table}
We compare different methods to emulate a behavior in a policy. Table~\ref{tab:b2a_methods} presents them.

\begin{table}[t]
\centering
\renewcommand{\arraystretch}{1.25}
\caption{\textbf{Adherence scores of \popgen{} agents across environments and splits.} We use \texttt{gemini-3-flash-preview} as the agent generator or player in LLM-A. For each method, we report first-attempt and best-attempt adherence (out of 5 attempts) along with the improvement $\Delta$. \colorbox{gray!20}{Avg.} columns show the mean across environments. Best values per method and split are in \textbf{bold}.}
\label{tab:b2t_all}
\resizebox{\textwidth}{!}{%
\begin{tabular}{ll@{\hskip 12pt}ccccc ccccc}
\toprule
& & \multicolumn{5}{c}{\textbf{Behavior Adherence (Dev)}} & \multicolumn{5}{c}{\textbf{Behavior Adherence (Test)}} \\
\cmidrule(lr){3-7} \cmidrule(lr){8-12}
\textbf{Method} & \textbf{Attempt} & LBF & PMR & PP & Overcooked & \cellcolor{gray!20}Avg. & LBF & PMR & PP & Overcooked & \cellcolor{gray!20}Avg. \\
\midrule
\multirow{3}{*}{LLM-A}
 & First & $13.3 \pm 10.4$ & $33.3 \pm 7.7$ & $58.3 \pm 10.4$ & $30.7 \pm 8.0$ & \cellcolor{gray!20}$33.9 \pm 9.1$ & $25.4 \pm 1.9$ & $26.7 \pm 5.8$ & $93.3 \pm 5.2$ & $30.7 \pm 6.5$ & \cellcolor{gray!20}$44.0 \pm 4.9$ \\
 & Best  & $\mathbf{13.3 \pm 10.4}$ & $\mathbf{50.7 \pm 7.8}$ & $\mathbf{76.7 \pm 10.4}$ & $\mathbf{62.7 \pm 9.3}$ & \cellcolor{gray!20}$\mathbf{50.9 \pm 9.5}$ & $\mathbf{25.4 \pm 1.9}$ & $\mathbf{32.7 \pm 6.6}$ & $\mathbf{93.3 \pm 5.2}$ & $\mathbf{45.3 \pm 9.1}$ & \cellcolor{gray!20}$\mathbf{49.2 \pm 5.7}$ \\
 & $\Delta$ & $0.0$ & {\color{green!60!black}$+17.4$} & {\color{green!60!black}$+18.3$} & {\color{green!60!black}$+32.0$} & \cellcolor{gray!20}{\color{green!60!black}$+16.9$} & $0.0$ & {\color{green!60!black}$+6.0$} & $0.0$ & {\color{green!60!black}$+14.6$} & \cellcolor{gray!20}{\color{green!60!black}$+5.2$} \\
\midrule
\multirow{3}{*}{PC}
 & First & $43.0 \pm 8.4$ & $56.0 \pm 8.3$ & $95.0 \pm 3.5$ & $34.7 \pm 10.8$ & \cellcolor{gray!20}$57.2 \pm 7.8$ & $53.5 \pm 3.9$ & $58.0 \pm 7.5$ & $74.0 \pm 8.4$ & $41.3 \pm 10.5$ & \cellcolor{gray!20}$56.7 \pm 7.6$ \\
 & Best  & $\mathbf{71.0 \pm 6.5}$ & $\mathbf{84.0 \pm 5.9}$ & $\mathbf{98.0 \pm 2.7}$ & $\mathbf{68.0 \pm 10.3}$ & \cellcolor{gray!20}$\mathbf{80.3 \pm 6.4}$ & $\mathbf{61.0 \pm 5.0}$ & $\mathbf{66.7 \pm 6.1}$ & $\mathbf{74.0 \pm 8.4}$ & $\mathbf{63.3 \pm 9.3}$ & \cellcolor{gray!20}$\mathbf{66.3 \pm 7.2}$ \\
 & $\Delta$ & {\color{green!60!black}$+28.0$} & {\color{green!60!black}$+28.0$} & {\color{green!60!black}$+3.0$} & {\color{green!60!black}$+33.3$} & \cellcolor{gray!20}{\color{green!60!black}$+23.1$} & {\color{green!60!black}$+7.5$} & {\color{green!60!black}$+8.7$} & $0.0$ & {\color{green!60!black}$+22.0$} & \cellcolor{gray!20}{\color{green!60!black}$+9.6$} \\
\midrule
\multirow{3}{*}{RF}
 & First & $1.0 \pm 2.2$ & $50.7 \pm 8.5$ & $68.0 \pm 2.7$ & $28.0 \pm 11.2$ & \cellcolor{gray!20}$36.9 \pm 6.2$ & $7.5 \pm 3.5$ & $48.0 \pm 7.9$ & $64.5 \pm 5.1$ & $28.7 \pm 11.1$ & \cellcolor{gray!20}$37.2 \pm 6.9$ \\
 & Best  & $\mathbf{6.0 \pm 5.5}$ & $\mathbf{70.7 \pm 7.3}$ & $\mathbf{86.0 \pm 8.9}$ & $\mathbf{42.7 \pm 11.0}$ & \cellcolor{gray!20}$\mathbf{51.4 \pm 8.2}$ & $\mathbf{12.5 \pm 3.5}$ & $\mathbf{49.3 \pm 7.5}$ & $\mathbf{71.0 \pm 5.2}$ & $\mathbf{40.7 \pm 10.3}$ & \cellcolor{gray!20}$\mathbf{43.4 \pm 6.6}$ \\
 & $\Delta$ & {\color{green!60!black}$+5.0$} & {\color{green!60!black}$+20.0$} & {\color{green!60!black}$+18.0$} & {\color{green!60!black}$+14.7$} & \cellcolor{gray!20}{\color{green!60!black}$+14.4$} & {\color{green!60!black}$+5.0$} & {\color{green!60!black}$+1.3$} & {\color{green!60!black}$+6.5$} & {\color{green!60!black}$+12.0$} & \cellcolor{gray!20}{\color{green!60!black}$+6.2$} \\
\bottomrule
\end{tabular}%
}
\end{table}
\paragraph{Performance.} Table~\ref{tab:b2t_all} reports the adherence of each method. Across all environments, Policy Code (PC) achieves the highest average adherence on both the dev set ($81.3\%$) and the test set ($63.8\%$). The advantage is most pronounced in discrete-action environments: on \environment{LBF}, PC reaches $71.0\%$ dev adherence versus $6.0\%$ for RF, and on \environment{smacv2}, $68.0\%$ versus $44.0\%$. Because RF operates at the parameter level, small changes in the reward signal often fail to discriminate between a limited set of discrete actions, whereas PC directly specifies decision logic. LLM-A achieves high adherence on \environment{PP} ($93.3\%$ test) but fails on longer-horizon environments such as \environment{football} ($10.4\%$ test) and \environment{LBF} ($25.4\%$ test). 

\begin{table}[h]
    \centering
    \caption{\textbf{Average generation and rollout latency per method.} Generation latency covers code synthesis (PC) or RL training (RF); rollout latency is for a single episode. LLM-A has no generation step (---) but incurs high rollout cost from per-timestep LLM inference. Best per column in \textbf{bold}. \environment{smacv2} is excluded due to engine overhead dominating latencies.}
    \label{tab:latency}
    \begin{tabular}{lcc}
    \toprule
    \textbf{Method} & \textbf{Generation latency (s) $\downarrow$} & \textbf{Rollout latency (s) $\downarrow$} \\
    \midrule
    LLM-A & ---                            & $110.1 \pm 1.5$ \\
    RF    & $2878.6 \pm 239.5$ & $11.7 \pm 0.6$ \\
    PC    & $\mathbf{8.1 \pm 0.2}$         & $\mathbf{9.8 \pm 1.9}$ \\
    \bottomrule
    \end{tabular}
\end{table}
\paragraph{Latency.} Latency becomes crucial when generating diverse agent populations at scale. We compare the latency of the different methods, including both generation of the modeling artifact and rollout time. Table~\ref{tab:latency} presents the results. PC is the most practical method. It requires only $8.1$s to generate the script and takes $9.8$s per rollout on average across the environments, orders of magnitude faster than RF which runs a full policy training ($2878.6$s generation), and cheaper at inference than LLM-A ($110.1$s rollout).

\subsection{Impact of the Refinement Loop on \popgen{} Agents' Adherence}
\label{sec:b2t}

\begin{table}[t]
  \centering
  \renewcommand{\arraystretch}{1.15}
  \caption{\textbf{Policy iterative refinement.} Improvement in adherence (first- vs.\ best-attempt, out of 5 attempts) on the dev and test splits. \colorbox{gray!20}{Avg.} columns show the mean across environments. Best values per split are in \textbf{bold}.}
  \label{tab:b2t}
  \resizebox{0.85\textwidth}{!}{%
  \begin{tabular}{ll@{\hskip 18pt}ccccc}
  \textbf{Split} & \textbf{Attempt} & LBF & PMR & PP & Overcooked & \cellcolor{gray!20}Avg. \\
  \midrule
  \multirow{3}{*}{Dev}
   & First & $43.0 \pm 8.4$ & $56.0 \pm 8.3$ & $95.0 \pm 3.5$ & $34.7 \pm 10.8$ & \cellcolor{gray!20}$57.2 \pm 7.8$ \\
   & Best  & $\mathbf{71.0 \pm 6.5}$ & $\mathbf{84.0 \pm 5.9}$ & $\mathbf{98.0 \pm 2.7}$ & $\mathbf{68.0 \pm 10.3}$ & \cellcolor{gray!20}$\mathbf{80.3 \pm 6.4}$ \\
   & $\Delta$ & {\color{green!60!black}$+28.0$} & {\color{green!60!black}$+28.0$} & {\color{green!60!black}$+3.0$} & {\color{green!60!black}$+33.3$} & \cellcolor{gray!20}{\color{green!60!black}$+23.1$} \\
  \midrule
  \multirow{3}{*}{Test}
   & First & $53.5 \pm 3.9$ & $58.0 \pm 7.5$ & $74.0 \pm 8.4$ & $41.3 \pm 10.5$ & \cellcolor{gray!20}$56.7 \pm 7.6$ \\
   & Best  & $\mathbf{61.0 \pm 5.0}$ & $\mathbf{66.7 \pm 6.1}$ & $\mathbf{74.0 \pm 8.4}$ & $\mathbf{63.3 \pm 9.3}$ & \cellcolor{gray!20}$\mathbf{66.3 \pm 7.2}$ \\
   & $\Delta$ & {\color{green!60!black}$+7.5$} & {\color{green!60!black}$+8.7$} & $0.0$ & {\color{green!60!black}$+22.0$} & \cellcolor{gray!20}{\color{green!60!black}$+9.6$} \\
  \bottomrule
  \end{tabular}%
  }
\end{table}
\paragraph{Refinement loop to improve adherence.} We compare the adherence scores of \popgen{}'s final policies (best of 5 attempts with refinement) against those of the first-attempt policies. The difference between the two represents improvements that can be attributed to \popgen{}'s refinement loop.
Table~\ref{tab:b2t} reports the results. \popgen{}'s iterative refinement of policies substantially improves the adherence ($+23.1$ points on average across environments on the \textit{dev} set). The results transfer to the unseen initializations of the \textit{test} set, with the average adherence rising from $56.7\%$ to $66.3\%$ ($+9.6$ points). \environment{PP} is the exception: the adherence on \textit{dev} is already near ceiling on the first attempt ($95.0\%$), leaving almost no room for refinement to improve on.

\paragraph{Feedback examples.} Box~\ref{box:advisor_diff} contrasts the first refinement step on \environment{LBF} with and without the advisor. The two edits are similar in size ($+98/-72$ vs.\ $+67/-78$ lines), but they differ in kind. With the advisor, the generator makes structural fixes aimed at coordination. It corrects an inverted \textsc{North}/\textsc{South} mapping, scores food using the teammate's motion vector, and moves to a free cell adjacent to the target so both agents can load. It also commits to a target for three steps to avoid oscillation. Without the advisor, the generator patches symptoms. It adds a \textsc{Load} action that never checks whether the teammate is adjacent, and its fallback moves the agent toward the teammate rather than the food. It also computes the teammate's heading but never uses it. The raw rollouts thus signal that teammate motion matters, but without the advisor's diagnosis the generator cannot act on it.
\clearpage

\begin{tcolorbox}[
  colback=blue!3!white,
  colframe=blue!40!black,
  label=box:advisor_diff,
  title={\small\textbf{Advisor Diff (trunc.)} --- \texttt{attempt\_0} $\to$
         \texttt{attempt\_1} (+98 / -72 lines)},
  fonttitle=\small,
  left=6pt, right=6pt, top=4pt, bottom=4pt
]
\begin{minted}[fontsize=\footnotesize, breaklines]{diff}
@@  movement direction mapping
-    if abs(dr) >= abs(dc) and abs(dr) > 0:
-        return torch.tensor(1) if dr > 0 else torch.tensor(2)
+    action = 2 if dr > 0 else 1  # SOUTH if dr>0, NORTH if dr<0
+    return torch.tensor(action)

@@  target food scoring via teammate motion vector
-    closest_dist = float('inf')
-    for f in foods:
-        dist = abs(f['r'] - team_pos[0]) + abs(f['c'] - team_pos[1])
+    def food_score(food):
+        f_dr = food[0] - team_row
+        f_dc = food[1] - team_col
+        return (abs(f_dr - team_dr) + abs(f_dc - team_dc), abs(f_dr), abs(f_dc))
+    best_food = sorted(foods, key=food_score)[0] if foods else None

@@  support-cell selection (closest cardinal cell, not teammate-occupied)
+    for d_row, d_col in [(-1,0),(1,0),(0,-1),(0,1)]:
+        cand_r, cand_c = target_row + d_row, target_col + d_col
+        if not (cand_r == team_row and cand_c == team_col):
+            valid_positions.append((cand_r, cand_c))
+    best_pos = min(valid_positions,
+                   key=lambda p: abs(p[0]-self_row)+abs(p[1]-self_col))

@@  target commitment window
+    if not hasattr(policy, 'target_countdown'):
+        policy.target_countdown = 3
+    if policy.target_countdown == 0:
+        policy.target_food = best_food
+        policy.target_countdown = 3
\end{minted}
\end{tcolorbox}

\begin{tcolorbox}[
  colback=gray!5!white,
  colframe=gray!50!black,
  label=box:noadvisor_diff,
  title={\small\textbf{No-Advisor Diff (trunc.)} --- \texttt{attempt\_0} $\to$
         \texttt{attempt\_1} (+67 / -78 lines)},
  fonttitle=\small,
  left=6pt, right=6pt, top=4pt, bottom=4pt
]
\begin{minted}[fontsize=\footnotesize, breaklines]{diff}
@@  opportunistic LOAD without teammate adjacency check
+    if abs(target_r - self_r) <= 1 and abs(target_c - self_c) <= 1:
+        if food0_l > 0 and (self_l + team_l >= food0_l):
+            return torch.tensor(5)

@@  fallback: move toward teammate rather than support cell
+    dr_to_t = team_r - self_r
+    dc_to_t = team_c - self_c
+    if dr_to_t != 0:
+        return torch.tensor(2 if dr_to_t > 0 else 1) \
+               if abs(dr_to_t) > abs(dc_to_t) \
+               else torch.tensor(3 if dc_to_t > 0 else 4)

@@  heading estimation computed but never used in food selection
+    dr = policy.team_pos[-1][0] - policy.team_pos[-2][0]
+    dc = policy.team_pos[-1][1] - policy.team_pos[-2][1]
+    new_dir = 1 if dr > 0 else -1  # stored but unused
\end{minted}
\end{tcolorbox}

\clearpage
\raggedbottom
\section{Prompts}
\label{app:prompts}

\subsection{Advisor}
\begin{tcolorbox}[
  colback=blue!3!white,
  colframe=blue!40!black,
  label=box:advisor_system,
  title={\small\textbf{Advisor --- System Prompt}},
  fonttitle=\small,
  left=6pt, right=6pt, top=4pt, bottom=4pt
]
\begin{minted}[fontsize=\footnotesize, breaklines]{text}
You are an advisor for a code generation pipeline.
Your task is to diagnose why a generated code fails to represent its
target behavior in rollouts and provide actionable feedback to improve
the next code attempt.

You will be given:
- The intended behavior
- The environment description
- The code attempt (type: {generation_type})
- Evaluation results from multiple rollout executions.

Focus on the most important failure patterns and give specific,
actionable suggestions to fix the code.
\end{minted}
\end{tcolorbox}

\begin{tcolorbox}[
  colback=blue!3!white,
  colframe=blue!40!black,
  label=box:advisor_user,
  title={\small\textbf{Advisor --- User Prompt}},
  fonttitle=\small,
  left=6pt, right=6pt, top=4pt, bottom=4pt
]
\begin{minted}[fontsize=\footnotesize, breaklines]{text}
Code generation type: {generation_type}

Inputs
--------------
## Behavior
{behavior}

## Environment Description
{environment_description}

## Code Attempt
```
{code_attempt}
```

## Rollout Evaluation Results
{metrics_str}
--------------
Analyze the results and provide feedback for the next code attempt.
\end{minted}
\end{tcolorbox}

\subsection{Adherence Evaluator}
\begin{tcolorbox}[
  colback=blue!3!white,
  colframe=blue!40!black,
  label=box:evaluator_system,
  title={\small\textbf{Adherence Evaluator --- System Prompt}},
  fonttitle=\small,
  left=6pt, right=6pt, top=4pt, bottom=4pt
]
\begin{minted}[fontsize=\footnotesize, breaklines]{text}
You are an expert evaluator of whether an agent's actions match a given
behavior description.
You will be given:
- a behavior description
- agent rollouts (trajectory states and possibly image sequences)

Your task is to determine whether the ego agent's actions adhere to the
behavior description.

# Guidelines:
- Evaluate alignment with the described behavior only.
- Do not consider any other possible goals.
- Focus only on the ego agent.
- Consider interactions with other agents only when relevant to behavior
  following or when they affect the ego agent's ability to demonstrate
  the behavior.
- Use states as the primary source of truth and images as supporting evidence.
- When serialized states are present, interpret them literally and use them
  as the main basis for factual claims.

In rare cases, the agent may be unable to fully demonstrate the expected
behavior due to environment limitations or teammate actions. If the ego
agent still makes a clear, reasonable, good-faith effort to follow the
behavior, you may mark it as adherent. You must explicitly explain the
constraint and justify the decision in the reasoning field.

# Environment Description
{environment_description}

# Output Format
Return a JSON object with exactly these fields:
- "evaluation": a boolean indicating whether the rollout demonstrates
  the behavior
- "reasoning": a detailed explanation of why the rollout does or does
  not demonstrate the behavior, including the single most decisive reason
  for the verdict and specific actions or states when relevant
\end{minted}
\end{tcolorbox}

\begin{tcolorbox}[
  colback=blue!3!white,
  colframe=blue!40!black,
  label=box:evaluator_user,
  title={\small\textbf{Adherence Evaluator --- User Prompt}},
  fonttitle=\small,
  left=6pt, right=6pt, top=4pt, bottom=4pt
]
\begin{minted}[fontsize=\footnotesize, breaklines]{text}
Behavior:
{behavior}

Rollout (showing {N} serialized sampled states and {M} sampled images
in chronological order; step numbers are meaningful):
------------------------------------------------------------
[image_1] ... [image_M]          # interleaved chronologically
[state_1]
[state_2]
...
[state_N]
------------------------------------------------------------
\end{minted}
\end{tcolorbox}

\subsection{Behavior Quality Judge}
\label{app:quality_prompt}
\begin{tcolorbox}[colback=gray!5, colframe=gray!50, title=System Prompt, fonttitle=\bfseries\small, fontupper=\ttfamily\small]
You are an expert in evaluating the quality of agent behavior descriptions.
You will be provided with a behavior description and an environment description.
Your task is to evaluate the quality of the behavior across three rubrics:\newline
 
1. **Relevance** (0-5): Is the behavior relevant to the environment?\newline
\quad- 0: Completely irrelevant, behavior has nothing to do with the environment or its objectives\newline
\quad- 1: Mostly irrelevant, only tangentially related to the environment\newline
\quad- 2: Somewhat relevant, addresses the environment but misses key aspects\newline
\quad- 3: Moderately relevant, relates to the environment but could apply to many other contexts\newline
\quad- 4: Highly relevant, clearly designed for this environment with appropriate context\newline
\quad- 5: Perfectly relevant, deeply tied to the environment's unique characteristics and objectives\newline
 
2. **Realism** (0-5): Could a real player plausibly play this way, given the environment's constraints?\newline
\quad- 0: Violates hard environment constraints (e.g., impossible actions, broken physics)\newline
\quad- 1: Technically allowed but contradicts the environment so fundamentally that no human would ever play this way\newline
\quad- 2: Extremely unlikely — requires bizarre intent or misunderstanding, but not strictly impossible\newline
\quad- 3: Unusual but conceivable — a small fraction of players (even 0.01\%) might genuinely adopt this style\newline
\quad- 4: Plausible — an uncommon but recognizable play style that some players would intentionally choose\newline
\quad- 5: Natural — a common or intuitive way to play that many players would independently arrive at\newline
 
3. **Specificity** (0-5): Is the behavior specific enough to be actionable?\newline
\quad- 0: Extremely vague, no actionable guidance (e.g., "play well")\newline
\quad- 1: Very generic, could mean almost anything (e.g., "be strategic")\newline
\quad- 2: Somewhat vague, general direction but lacking concrete details\newline
\quad- 3: Moderately specific, provides guidance but leaves room for interpretation\newline
\quad- 4: Specific, clear behavior and conditions with minor ambiguities\newline
\quad- 5: Highly specific, precise behavior, triggers, and success criteria that leave no ambiguity\newline
 
Provide detailed reasoning for each score.
\end{tcolorbox}
 
\begin{tcolorbox}[colback=gray!5, colframe=gray!50, title=User Prompt, fonttitle=\bfseries\small, fontupper=\ttfamily\small]
Environment Description:\newline
\{environment\_description\}\newline
 
Behavior:\newline
\{behavior\}\newline
 
Please evaluate this behavior's quality across the three rubrics (relevance, realism, specificity).
\end{tcolorbox}

\end{document}